%% file: main.tex
\documentclass[10pt,twocolumn,letterpaper]{article}

\usepackage{multirow}
\usepackage{graphicx}
\usepackage{cvpr}
\usepackage{times}
\usepackage{epsfig}
\usepackage{graphicx}
\usepackage{amsmath}
\usepackage{amssymb}
\usepackage{booktabs}
\usepackage{bm}
\usepackage{dsfont}
\usepackage{subcaption}

\usepackage{pifont}
\usepackage{parskip}
\usepackage[nocompress]{cite}
\usepackage[pagebackref=true,breaklinks=true,colorlinks,bookmarks=false]{hyperref}
\usepackage{cleveref}
\usepackage{textcomp}
\usepackage{dblfloatfix}    %

\usepackage{color}
\usepackage{wrapfig,booktabs}
\usepackage{caption}
\usepackage{booktabs}
\usepackage[table,xcdraw]{xcolor}

\usepackage{tabularx}
\usepackage{multicol}
\usepackage{enumitem}
\usepackage{calc}
\usepackage{dblfloatfix}    %

\newcommand{\cmark}{\ding{51}}%
\newcommand{\xmark}{\ding{55}}%

\newcommand{\bx}{\bm{x}}

\newcommand{\bg}{\bm{g}}
\newcommand{\bw}{\bm{w}}
\newcommand{\by}{\bm{y}}

\cvprfinalcopy %

\begin{document}

\title{Adaptive Text Recognition through Visual Matching}

\author{Chuhan Zhang\\
University of Oxford\\
{\tt\small czhang@robots.ox.ac.uk}
\and
Ankush Gupta\\
DeepMind, London\\
{\tt\small ankushgupta@google.com}
\and
Andrew Zisserman\\
University of Oxford\\
{\tt\small az@robots.ox.ac.uk}
}

\maketitle

\begin{abstract}
   \input{sections/abstract}
\end{abstract}

\input{figs/teaser}

\input{sections/Intro.tex}
\input{sections/related_work.tex}

\input{figs/arch}
\input{sections/method.tex}

\input{tables/arch_encoder}
\input{sections/implementation.tex}

\input{sections/exps.tex}
\input{sections/conclusion.tex}
{\small
	\bibliographystyle{ieee_fullname}
	\bibliography{bib/shortstrings,bib/refs,bib/vgg_local,bib/vgg_other,bib/egbib}
}

\appendix

\onecolumn
\noindent {\huge\textbf{Appendix}}
\\
\\

\input{supp/supp.tex}

\end{document}

%% file: sections/abstract.tex
This work addresses the problems of generalization and flexibility for text 
recognition in documents. We introduce a new model that exploits the repetitive
nature of characters in languages, and decouples the visual decoding and 
linguistic modelling stages through intermediate representations in the form of 
similarity maps. By doing this, we turn text recognition into a 
visual matching problem, thereby achieving generalization in appearance
and flexibility in classes.

We evaluate the model on both synthetic and real datasets across different 
languages and alphabets, and show that it can handle challenges that 
traditional architectures are unable to solve without expensive re-training, 
including: (i) it can change  the number of classes simply by changing the exemplars; 
and (ii) it can generalize  to novel languages and characters (not in the training data)
simply by providing a new glyph exemplar  set. 
In essence, it is able to carry out one-shot sequence recognition.
We also demonstrate that the model can generalize to unseen fonts without requiring
new exemplars from them.

Code, data, and model checkpoints are available at:\\
{\small\url{http://www.robots.ox.ac.uk/~vgg/research/FontAdaptor20/}}.

%% file: figs/teaser.tex
\begin{figure}[t] 
  \begin{center}
     \includegraphics[width=0.95\linewidth]{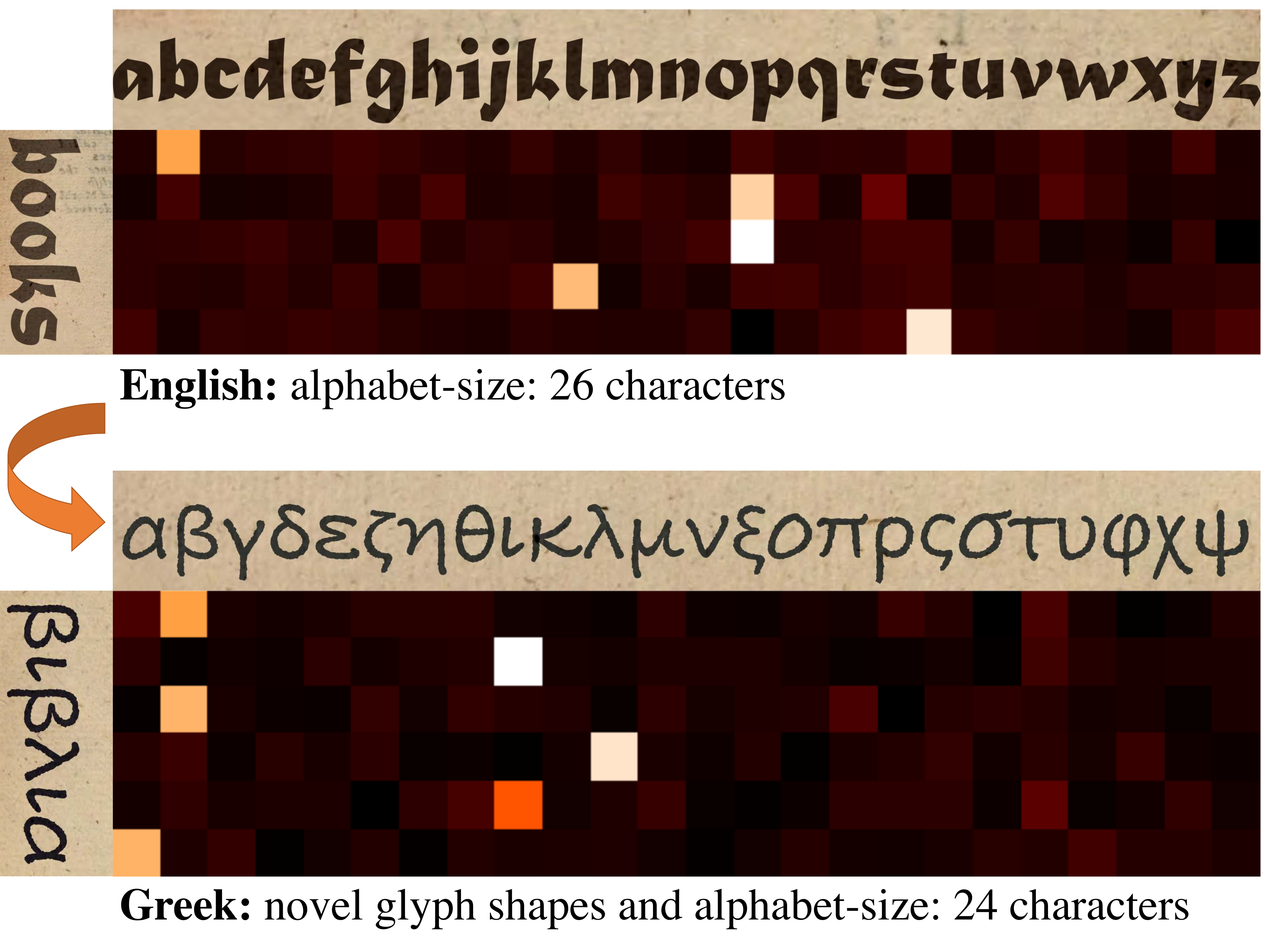}
  \end{center}
   \caption{\textbf{Visual matching for text recognition.}
     Current text recognition models learn discriminative features specific to character shapes (\emph{glyphs}) from a pre-defined (fixed) alphabet. We train our model instead to establish \emph{visual similarity} between given character glyphs \textbf{(top)} and the text-line image to be recognized \textbf{(left)}. This makes the model highly adaptable to unseen glyphs, new alphabets (different languages) and extensible to novel character classes, \eg, English $\rightarrow$ Greek, \emph{without} further training.
     Brighter colors correspond to higher visual similarity.}
  \label{fig:teaser}
\end{figure}

%% file: sections/intro.tex
\section{Introduction}\label{sec:intro}

Our objective in this work is {\em generalization} and {\em flexibility} in text
recognition. Modern text recognition methods~\cite{shi2018aster,cheng2018aon,
Lee16,baek2019wrong} achieve excellent performance in many cases, but 
generalization to unseen data, \ie, novel fonts and new languages, 
either requires large amounts of data for primary training or expensive 
fine-tuning for each new case.  

The text recognition problem is to map an image of a line of text $\bx$ into the
corresponding sequence of characters $\by = (y_1,y_2,\hdots,y_k)$, where $k$ is 
the length of the string and $y_i
\in \mathcal{A}$ are characters in alphabet $\mathcal{A}$ (\eg, 
\texttt{\{a,b,$\hdots$,z,<space>\}}). Current deep learning based 
methods~\cite{shi2018aster,cheng2018aon,Lee16} cast this in the encoder-decoder 
framework~\cite{sutskever2014sequence,cho2014learning}, where
first the text-line image is encoded through a visual ConvNet~\cite{Lecun89},
followed by a recurrent neural network decoder, with alignment between 
the visual features and text achieved either through 
attention~\cite{bahdanau2014neural} or Connectionist Temporal Classification 
(CTC)~\cite{graves2005framewise}.

\noindent\textbf{Impediments to generalization.} The conventional methods for 
text recognition train the visual encoder and the sequence decoder modules in an 
end-to-end manner. While this is desirable for optimal co-adaptation, it 
induces monolithic representations which confound visual and linguistic 
functions. Consequently, these methods suffer from the following limitations:
(1)~Discriminative recognition models specialize to fonts and textures in
the training set, hence generalize poorly to novel visual styles.
(2)~The decoder discriminates over a fixed alphabet/number of 
characters.
(3)~The encoder and decoder are tied to each other, hence are not 
inter-operable across encoders for new visual styles or decoders for new languages.
Therefore, current text recognition methods generalize poorly and 
require re-initialization or fine-tuning for new alphabets and languages. 
Further, fine-tuning typically requires new training data for the target domain
and does not overcome these inherent limitations.

\noindent\textbf{Recognition by matching.} Our method is based on a key 
insight: text is a sequence of repetitions of a finite number of 
discrete entities. The repeated entities are \emph{characters} in a text string,
and \emph{glyphs}, \ie, visual representations of characters/symbols, in a 
text-line image. We re-formulate the text recognition problem as one of 
\emph{visual matching}. We assume access to \emph{glyph exemplars} (\ie, cropped
images of characters), and task the visual encoder to localize these repeated 
glyphs in the given text-line image. The output of the visual encoder is a 
\emph{similarity map} which encodes the visual similarity of each spatial 
location in the text-line to each glyph in the alphabet as shown in 
\Cref{fig:teaser}. The decoder ingests this similarity map to infer the most 
probable string. \Cref{f:arch} summarizes the proposed method.

\noindent\textbf{Overcoming limitations.} The proposed model overcomes the above
mentioned limitations as follows:
(1)~Training the encoder for \emph{visual matching} relieves it from 
	learning specific visual styles (fonts, colors \etc) from the training data, 
	improving generalization over novel visual styles.
(2)~The similarity map is agnostic to the number of different glyphs, hence 
	the model generalizes to novel alphabets (different number of characters).
(3)~The similarity map is also agnostic to visual styles, and acts as
	an interpretable interface between the visual encoder and the decoder, 
	thereby disentangling the two.

\noindent\textbf{Contributions.} Our main contributions are threefold.
	First, we propose a novel network design for text recognition aimed at 
	generalization. We exploit the repetition of glyphs in language, and build 
	this similarity between units into our architecture. The model is described
	in~\Cref{sec:model,s:impl}.
	Second, we show that the model outperforms state-of-the-art 
	methods in recognition of novel fonts unseen during training~(\Cref{s:exps}).
	Third, the model can be applied to novel languages without
	expensive fine-tuning at test time; it is only necessary to supply glyph 
	exemplars for the new font set. These include languages/alphabets with 
	different number of characters, and novel styles \eg, characters with accents 
	or historical characters 
	\textit{`{\fontfamily{jkpkvos}\selectfont s}'}~(also in \Cref{s:exps}).

Although we demonstrate our model for \emph{document OCR} where a consistent 
visual style of glyphs spans the entire document, the method is applicable to
scene-text/text-in-the-wild (\eg, SVT~\cite{Wang10b}, 
ICDAR~\cite{Karatzas13,Karatzas15} datasets) where each instance has a unique 
visual style (results in \Cref{sec:scene}).

%% file: sections/related_work.tex
\section{Related Work}\label{s:rel_work}

\noindent\textbf{Few-shot recognition.}
Adapting model behavior based on class exemplars has been explored for
few-shot object recognition. Current popular few-shot classification 
methods, \eg, Prototypical Nets~\cite{snell2017prototypical}, Matching 
Nets\cite{vinyals2016matching}, Relation Nets~\cite{sung2018learning}, and 
MAML~\cite{finn2017model}, have been applied only to recognition of \emph{single
instances}. Our work addresses the unique challenges associated with 
one-shot classification of \emph{multiple instances in sequences}.
To the best of our knowledge this is the first work to address one-shot 
\emph{sequence recognition}. 
We discuss these challenges and the proposed architectural innovations 
in~\Cref{s:one-shot-seq}.
A relevant work is from Cao~\etal~\cite{cao2020few} which tackles few-shot 
video classification, but similar to few-shot object recognition methods,
they classify the whole video as a \emph{single} instance. 

\noindent\textbf{Text recognition.}
Recognizing text in images is a classic problem in pattern recognition.
Early successful applications were in reading handwritten documents~\cite{Lecun89,bunke04},
and document optical character recognition~(OCR)~\cite{tesseract-ocr}.
The OCR industry standard---\emph{Tesseract}~\cite{tesseract-ocr}---employs 
specialized training data for each supported language/alphabet.\footnote{
Tesseract's specialized training data for 103 languages:
\\\url{https://github.com/tesseract-ocr/tesseract/wiki/Data-Files}} Our model 
enables rapid adaptation to novel visual styles and alphabets and does not
require such expensive fine-tuning/specialization.
More recently, interest has been focussed towards text in natural images.
Current methods either directly classify word-level images~\cite{Jaderberg14c},
or take an encoder-decoder approach~\cite{sutskever2014sequence,cho2014learning}.
The text-image is encoded through a ConvNet, followed by 
bidirectional-LSTMs for context aggregation. 
The image features are then aligned with string labels either using 
Connectionist Temporal Classification (CTC)~\cite{graves2005framewise,Su14,
shi2016end,he2016reading} or through attention~\cite{bahdanau2014neural,
Lee16,Cheng17,cheng2018aon,shi2016robust}. Recognizing irregularly shaped 
text has garnered recent interest which has seen a resurgence of dense 
character-based segmentation and classification 
methods~\cite{feng2019textdragon,Lyu_2018_ECCV}.
Irregular text is rectified before feature extraction either using geometric 
transformations~\cite{Zhan_2019_CVPR,liu2018char,shi2018aster,shi2016robust} 
or by re-generating the text image in canonical fonts and 
colors~\cite{Liu_2018_ECCV}. Recently, Baek~\etal~\cite{baek2019wrong} present a
thorough evaluation of text recognition methods, unifying them in a four-stage 
framework---input transformation, feature extraction, sequence modeling, and 
string prediction.

%% file: figs/arch.tex
\begin{figure*}[t]
  	\begin{subfigure}[t]{0.75\linewidth}
  		     \includegraphics[width=\linewidth]{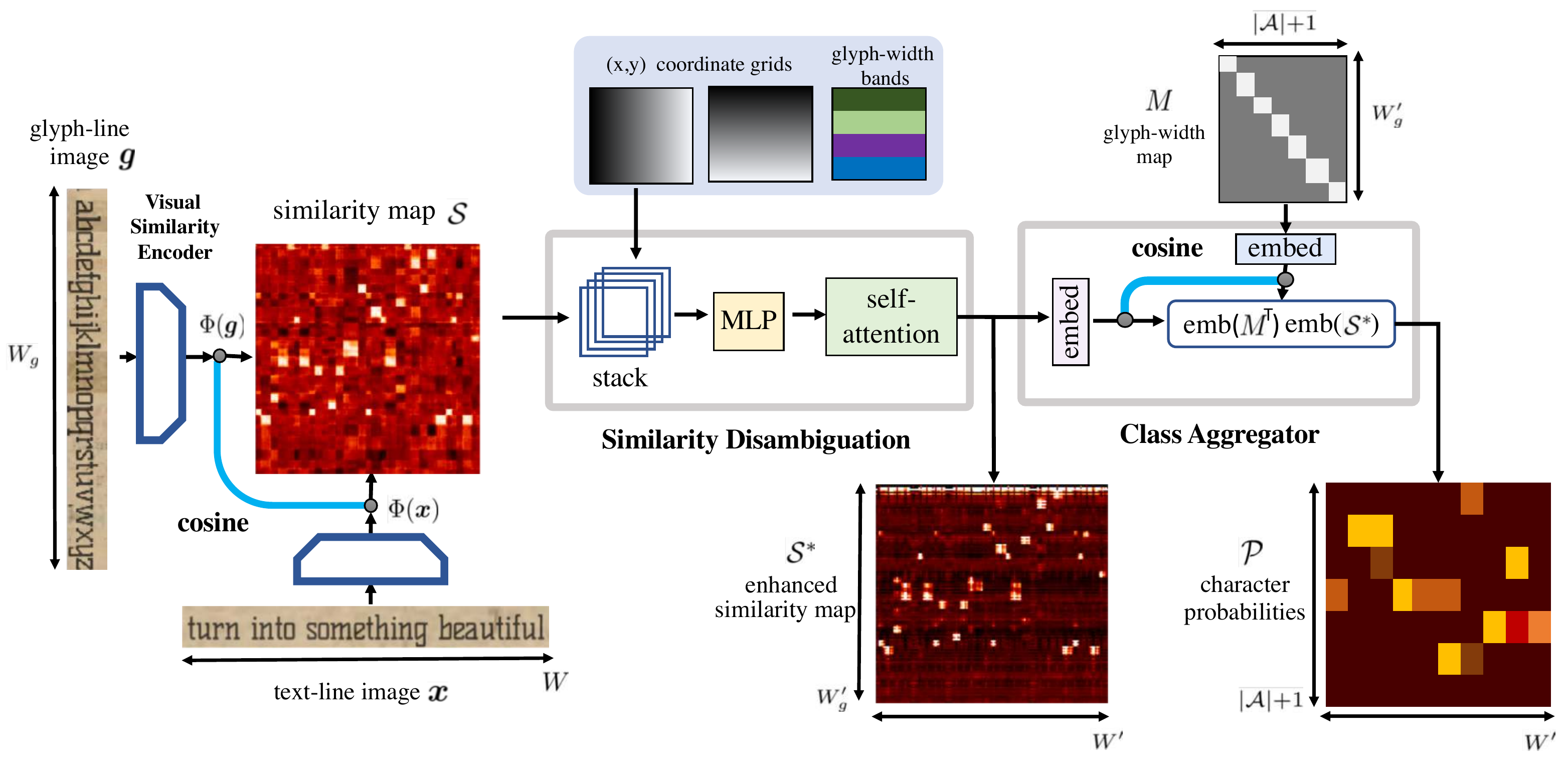}
  	\end{subfigure}
  	\begin{subfigure}[t]{0.23\linewidth}
			\includegraphics[width=\linewidth]{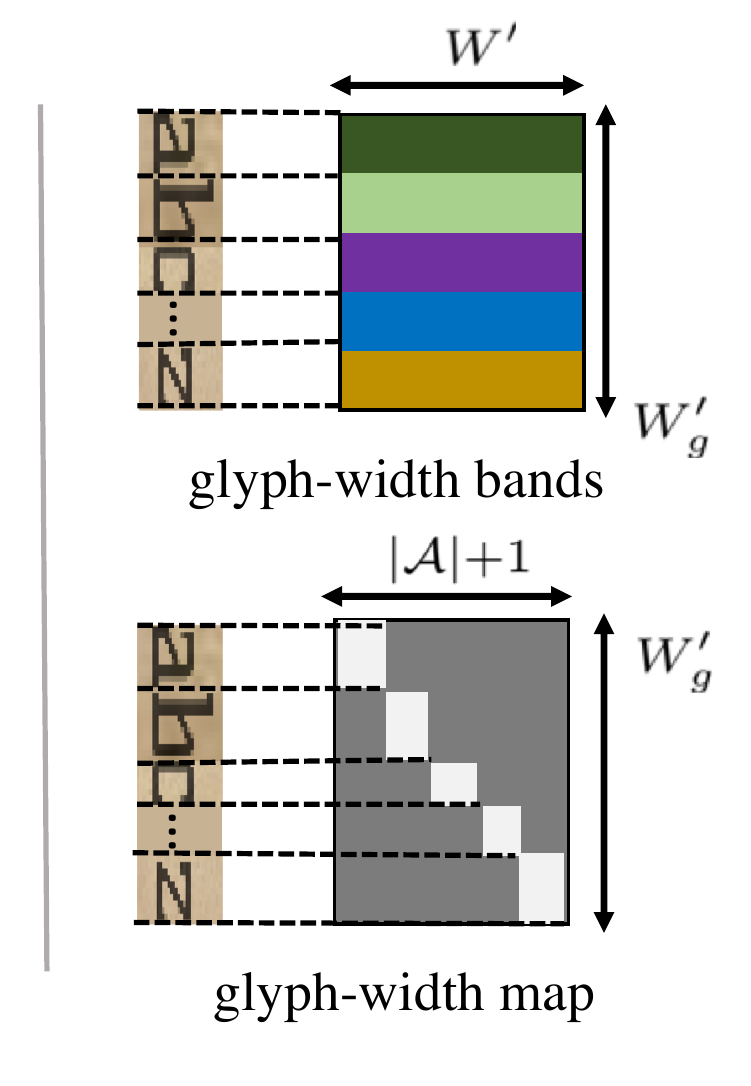}
		\end{subfigure}
		\caption{\textbf{Architecture for adaptive visual matching.} 
		 We cast the problem of text recognition as one of visual matching of glyph 
		 exemplars in the given text-line image. \textbf{Left}: Overview of the 
		 architecture. The visual encoder $\Phi$ embeds the glyph-line $\bg$ and 
		 text-line $\bx$ images and produces a similarity map $\mathcal{S}$, which 
		 scores the similarity of each glyph against each position along the 
		 text-line. Then, ambiguities in (potentially) imperfect visual matching are
		 resolved to produce the enhanced similarity map $\mathcal{S}^*$. Finally, 
		 similarity scores are aggregated to output class probabilities 
		 $\mathcal{P}$ using the ground-truth glyph width contained 
		 in~$\mathcal{M}$. 
		 \textbf{Right}: Illustration of how glyph-widths are encoded into the 
		 model. 
		 The glyph-width bands {\em (top)} have the same height as the width of 
		 their corresponding glyph exemplars, and their scalar values are the glyph 
		 widths in pixels.
		 The glyph-width map {\em (bottom)} is a binary 
		 matrix with a column for each character in the alphabet $\mathcal{A}$;
		 the columns indicate the extent of glyphs in the glyph-line image
		 by setting the corresponding rows to a non-zero value (${=}1$).}
  \label{f:arch}
  \vspace{-3mm}
\end{figure*}

%% file: sections/method.tex
\section{Model Architecture}\label{sec:model}
Our model recognizes a given text-line image by localizing glyph exemplars in it 
through visual matching. It takes both the text-line image and an alphabet image
containing a set of exemplars as input, and predicts a sequence of probabilities
over $N$ classes as output, where $N$ is equal to the number of exemplars given
in the alphabet image.
For inference, a glyph-line image is assembled from the individual 
character glyphs of a reference font simply by concatenating them side-by-side, and text-lines in that font can then be read.

The model has two main components:
(1)~a visual similarity encoder (\Cref{s:encoder}) which outputs a similarity map encoding the similarity of each glyph in the text-line image, and (2)~an alphabet agnostic decoder (\Cref{s:decoder}) which ingests this similarity map to infer the most probable string. In \Cref{s:str-pred} we give details for the training objective. \Cref{f:arch} gives a concise schematic of the model.

\subsection{Visual Similarity Encoder}\label{s:encoder}
The visual similarity encoder is provided with a set of glyphs for the target alphabet, and tasked to localize these glyphs in the input text-line image to be recognized. It first embeds the text-line and glyphs using a shared visual encoder $\Phi$ and outputs a \emph{similarity map} $\mathcal{S}$ which computes the visual similarity between all locations in the text-line against all locations in every glyph in the alphabet.

Mathematically, let $\bx \in \mathbb{R}^{H \times W \times C}$ be the text-line 
image, with height $H$, width $W$ and $C$ channels. Let the glyphs be 
$\{g_i\}_{i=1}^{i=|\mathcal{A}|}$, $g_i \in  \mathbb{R}^{H \times W_i \times C}$, 
where $\mathcal{A}$ is the exemplar set -- the alphabet plus the blank space and image padding, and $W_i$ is the width of the $i^{th}$ 
glyph. The glyphs are stacked along the width to form a \emph{glyph-line image} 
$\bg \in \mathbb{R}^{H \times W_g \times C}$.
Embeddings are obtained using the visual encoder $\Phi$ for both the text-line 
$\Phi(\bx) \in \mathbb{R}^{1 \times W' \times D}$ and the glyph-line 
$\Phi(\bg) \in \mathbb{R}^{1 \times W_g' \times D}$, where $D$ is the embedding
dimensionality. The output widths are downsampled by the network stride 
$s$ (\ie, $W' = \frac{W}{s}$). Finally, each spatial location along the width in 
the glyph-line image is scored against the every location in the text-line image
to obtain the similarity map $\mathcal{S} \in [-1, 1]^{W_g'\times W'}$:
\begin{equation}
    S_{ij} = \langle \Phi(\bg)_{i}, \Phi(\bx)_j \rangle = \frac{ \Phi(\bg)_{i}^{T} \Phi(\bx)_{j}}
    {||\Phi(\bg)_i||\cdot||\Phi(\bx)_j||} \label{e:sim} 
\end{equation}
where score is the cosine similarity, and $i\in\{1,\hdots,W_g'\}$, $j\in\{1,\hdots,W'\}$.

\subsection{Alphabet Agnostic Decoder}\label{s:decoder}
The alphabet agnostic decoder discretizes the similarity maps into probabilities for each glyph in the exemplars for all spatial locations along the width of the text-line image.
Concretely, given the visual similarity map $\mathcal{S} \in \mathbb{R}^{W_g'\times W'}$ it outputs logits over the glyph exemplars for each location in the text-line: $\mathcal{P}\in \mathbb{R}^{|A|\times W'}$, $\mathcal{P}_{ij} = \log p(y_i | \bx_j)$, where $\bx_j$ is the $j^{th}$ column in text-line image (modulo encoder stride) and $y_i$ is the $i^{th}$ exemplar in $\mathcal{A}$.

A simple implementation would predict the $\operatorname{argmax}$ or sum of the 
similarity scores aggregated over the extent of each glyph in the similarity map. 
However, this naive strategy does not overcome ambiguities in similarities or 
produce smooth/consistent character predictions. 
Hence, we proceed in two steps: first, \textbf{similarity disambiguation} 
resolves ambiguities over the glyphs in the alphabet producing an enhanced 
similarity map ($\mathcal{S}^*$) by taking into account the glyph widths and 
position in the line image, and second, \textbf{class aggregator} computes 
glyph class probabilities by aggregating the scores inside the spatial 
extent of each glyph in $\mathcal{S}^*$. We detail the two steps next; the 
significance of each component is established empirically in 
\Cref{s:exp-ablation}.

\noindent\textbf{Similarity disambiguation.} 
An ideal similarity map would have square regions of high-similarity.
This is because the width of a character in the glyph and text-line images
will be the same. Hence, we encode glyph widths
along with local $x$, $y$ coordinates using a small MLP into the similarity map.
The input to the MLP at each location is the similarity map value 
$\mathcal{S}$ stacked with:  (1) two channels of $x$, $y$ coordinates (normalized to $[0,1]$), 
and (2) \emph{glyph-width bands} $\mathcal{G}$: $\mathcal{G} = \bw_g \mathds{1}^T$, 
where $\bw_g \in \mathbb{R}^{W_g'}$ is a vector of glyph widths 
in pixels; see \Cref{f:arch} for an illustration.
For disambiguation, we use a self-attention module~\cite{vaswani2017attention} which attends over columns of $\mathcal{S}$ and outputs the final enhanced similarity map $\mathcal{S}^*$ of the same size as $\mathcal{S}$.

\noindent\textbf{Class aggregator.} The class aggregator $\Lambda$ maps the similarity map to logits over the alphabet along the horizontal dimension in the text-line image: $\Lambda: \mathbb{R}^{W_g'\times W'} \mapsto \mathbb{R}^{|A|\times W'}$, $\mathcal{S}^* \mapsto \mathcal{P}$.
This mapping can be achieved by multiplication through a matrix $M \in \mathbb{R}^{|\mathcal{A}|\times W_g'}$ which aggregates (sums) the scores in the span of each glyph: $\mathcal{P} = M\mathcal{S}^*$, such that
$M =  [m_1, m_2, \dots, m_{|\mathcal{A}|}]^T$ and $m_i \in \{0,1\}^{W_g'} = [0,\ldots,0, 1,\ldots,1,0,\ldots,0]$ where the non-zero values correspond to the span of the $i^{th}$ glyph in the glyph-line image. 

In practice, we first embed columns of $\mathcal{S}^*$ and $M^T$ independently using learnt linear embeddings. The embeddings are $\ell_2$-normalized before the matrix product (equivalent to cosine similarity). We also expand the classes to add an additional ``boundary'' class (for CTC) using a learnt $m_{|\mathcal{A}|+1}$. Since, the decoder is agnostic to the number of characters in the alphabet, it generalizes to novel alphabets.

\subsection{Training Loss}\label{s:str-pred}
The dense per-pixel decoder logits over the glyph exemplars $\mathcal{P}$ are supervised using the CTC loss~\cite{graves2006connectionist}~($\mathcal{L}_{CTC}$) to align the predictions with the output label. We also supervise the similarity map output of the visual encoder $\mathcal{S}$ using an auxiliary cross-entropy loss ($\mathcal{L}_{sim}$) at each location. We use ground-truth character bounding-boxes for determining the spatial span of each character. The overall training objective is the following two-part loss,
\begin{align}
\mathcal{L}_{pred} &= \mathcal{L}_{CTC}\left(\operatorname{SoftMax}(\mathcal{P}), \by_{gt})\right) \\
        \mathcal{L}_{sim} &= - \sum_{ij} \log(\operatorname{SoftMax}(S_{y_ij})) \label{e:loss_sim}\\
        \mathcal{L}_{total} &=  \mathcal{L}_{pred} + \lambda\mathcal{L}_{sim} \label{e:loss_tot}
\end{align}
where, $\operatorname{SoftMax}(\cdot)$ normalization is over the glyph exemplars (rows), $\by_{gt}$ is the string label, and $y_i$ is the ground-truth character associated with the $i^{th}$ position in the glyph-line image.
The model is insensitive to the value of $\lambda$ within a reasonable range (see \Cref{s:loss_ablation}), and we use 
$\lambda = 1$ for a good balance of losses.

\subsection{Discussion: One-shot Sequence Recognition}\label{s:one-shot-seq}
Our approach can be summarized as a method for one-shot sequence recognition.
Note, existing few-shot methods~\cite{vinyals2016matching,snell2017prototypical,
jia2016dynamic,sung2018learning} are not directly applicable to this problem of 
one-shot sequence recognition, as they focus on classification of the whole of 
the input (\eg an image) as a single instance. Hence, these cannot address the following
unique challenges associated with (text) sequences:
{(1)~segmentation} of the imaged text sequence  into characters of different widths;
{(2)~respecting language-model/sequence-regularity} in the output.
We develop a novel neural architectural solutions for the above, namely: 
(1) A neural architecture with {\em explicit reasoning over similarity 
maps} for decoding sequences. The similarity maps are key for 
{\em generalization} at both ends---novel fonts/visual styles and new 
alphabets/languages respectively.
(2) Glyph width aware {\em similarity disambiguation}, which identifies 
contiguous square blocks in noisy similarity maps from novel data. This is 
critical for robustness against imprecise visual matching.
(3) {\em Class aggregator}, aggregates similarity scores
over the reference width-spans of the glyphs to produce character logit scores
over the glyph exemplars. It operates over a variable number of characters/classes 
and glyph-widths.
The importance of each of these components is established in the ablation 
experiments in \Cref{s:exp-ablation}.%

%% file: tables/arch_encoder.tex
\setlength\intextsep{5pt}
\begin{table}

\setlength{\tabcolsep}{6pt}
\resizebox{\linewidth}{!}{%
  \begin{tabular}{@{}lcrclcrcl@{}}
    \toprule
    \multicolumn{1}{c}{\multirow{2}{*}{layer}} & \multirow{2}{*}{kernel}           & \multicolumn{3}{c}{channels}           & \multirow{2}{*}{pooling}          & \multicolumn{3}{c}{output size}  \\
    \multicolumn{1}{c}{}                       &                                   & \multicolumn{3}{c}{in / out}           &                                   & \multicolumn{3}{c}{H${\times}$W} \\ \midrule
    \multicolumn{1}{l|}{conv1}                 & \multicolumn{1}{c|}{$3{\times}3$} & 1      & /  & \multicolumn{1}{l|}{64}  & \multicolumn{1}{c|}{max = (2, 2)} & 16      & ${\times}$    & W/2    \\
    \multicolumn{1}{l|}{resBlock1}             & \multicolumn{1}{c|}{$3{\times}3$} & 64     & /  & \multicolumn{1}{l|}{64}  & \multicolumn{1}{c|}{max = (1, 2)} & 8       & ${\times}$    & W/2    \\
    \multicolumn{1}{l|}{resBlock2}             & \multicolumn{1}{c|}{$3{\times}3$} & 64     & /  & \multicolumn{1}{l|}{128} & \multicolumn{1}{c|}{max = (2, 2)} & 4       & ${\times}$    & W/4    \\
    \multicolumn{1}{l|}{upsample}              & \multicolumn{1}{c|}{--}           &        & -- & \multicolumn{1}{l|}{}    & \multicolumn{1}{c|}{(2, 2)}       & 8       & ${\times}$    & W/2    \\
    \multicolumn{1}{l|}{skip}                  & \multicolumn{1}{c|}{$3{\times}3$} & 128+64 & /  & \multicolumn{1}{l|}{128} & \multicolumn{1}{c|}{--}           & 8       & ${\times}$    & W/2    \\
    \multicolumn{1}{l|}{pool}                  & \multicolumn{1}{c|}{--}           &        & -- & \multicolumn{1}{l|}{}    & \multicolumn{1}{c|}{avg = (2, 1)} & 4       & ${\times}$    & W/2    \\
    \multicolumn{1}{l|}{conv2}                 & \multicolumn{1}{c|}{$1{\times}1$} & 128    & /  & \multicolumn{1}{l|}{64}  & \multicolumn{1}{c|}{--}           & 4       & ${\times}$    & W/2    \\
    \multicolumn{1}{l|}{reshape}               & \multicolumn{1}{c|}{--}           & 64     & /  & \multicolumn{1}{l|}{256} & \multicolumn{1}{c|}{--}           & 1       & ${\times}$    & W/2    \\ \bottomrule
    \end{tabular}}
  \caption{\textbf{Visual encoder architecture}~(\Cref{s:encoder,s:impl-encoder}). The input is an image of size $32{\times}W{\times}1$ (height${\times}$width${\times}$channels).}\label{table:encoder}
  \end{table}

%% file: sections/implementation.tex
\section{Implementation details}\label{s:impl}
The architectures of the visual similarity encoder and the alphabet agnostic 
decoder are described in \Cref{s:impl-encoder} and \Cref{s:impl-decoder} respectively,
followed by training set up in \Cref{s:impl-train-opt}.

\subsection{Visual Similarity Encoder}\label{s:impl-encoder}
The visual similarity encoder ($\Phi$) encodes both the text-line ($\bx$)
and glyph-line ($\bg$) images into feature maps.
The inputs of height 32 pixels, width $W$ and 
1 channel (grayscale images) are encoded into a tensor of size 
$1{\times}\frac{W}{2}{\times}256$.
The glyph-line image's width is held fixed to a constant $W_g=720$~px: if $\sum_{i=1}^{i=|\mathcal{A}|} W_i < W_g$ the image is padded at the end using the \texttt{<space>} glyph, otherwise the image is downsampled bilinearly to a width of $W_g=720$~px.
The text-line image's input width is free (after resizing to a height of 32 proportionally).
The encoder is implemented as a U-Net~\cite{ronneberger2015u} with 
two residual blocks~\cite{He15}; detailed architecture in \Cref{table:encoder}.
The visual similarity map ($\mathcal{S}$) is obtained by taking the cosine distance 
between all locations along the width of the encoded features
from text-line $\Phi(\bx)$ and glyph-line $\Phi(\bg)$ images.

\subsection{Alphabet Agnostic Decoder}\label{s:impl-decoder}
\noindent\textbf{Similarity disambiguation.} We use the self-attention based
\emph{Transformer} model \cite{vaswani2017attention} with three layers with
four attention heads of vector  dimension 360 each.
The input to this module is the similarity map $\mathcal{S}$ stacked with
with local positions ($x$, $y$) and glyph widths, which are then encoded 
through a three-layer ($4{\times}16$, $16{\times}32$, $32{\times}1$) MLP with 
ReLU non-linearity~\cite{Nair10}.

\noindent\textbf{Class aggregator.}
The columns of $\mathcal{S}^*$ and glyph width templates (refer to \Cref{s:decoder}) 
are embedded independently using linear embeddings of size $W_g'{\times}W_g'$, 
where $W_g'= \frac{W_g}{s} = \frac{720}{2} = 360$ ($s =$ encoder stride).

\noindent\textbf{Inference.} 
We decode greedily at  inference, as is common after training  with CTC loss.
  No additional language model (LM) is used, except in Experiment 
  VS-3~(\Cref{s:exp-vs3}), where a 6-gram LM learnt from over 10M sentences from
  the WMT News Crawl (2015) English corpus~\cite{wmt2015} is combined with 
  the model output with beam-search using the algorithm 
  in~\cite{maas2015lexicon} (parameters: $\alpha{=}1.0$, $\beta{=}2.0$, 
  beam-width${=}15$).

\subsection{Training and Optimization}\label{s:impl-train-opt}

The entire model is trained end-to-end by minimizing 
the training objective~\Cref{e:loss_tot}.
We use online data augmentation on both the text-line and glyph
images, specifically random translation, crops, contrast, and blur.
All parameters, for both ours and SotA models, are initialized with random weights.
We use the Adam optimizer~\cite{kingma2014adam} with a constant learning rate of 0.001, 
a batch size of 12 and train until validation accuracy saturates (typically 100k iterations)
on a single Nvidia Tesla P40 GPU. The models are implemented in PyTorch~\cite{paszke2017automatic}.

%% file: sections/exps.tex
\section{Experiments}\label{s:exps}
\input{graphs/split}

We compare against state-of-the-art text-recognition models for generalization 
to novel fonts and languages. We first describe the models 
used for comparisons (\Cref{s:exp-baseline}), then datasets and evaluation metrics 
(\Cref{s:exp-data}), followed by an overview of the experiments 
(\Cref{s:exp-overview}), and a thorough component analysis of the model 
architecture (\Cref{s:exp-ablation}). Finally, we present the results
 (\Cref{s:exp-results}) of all the experiments. 

\subsection{State-of-the-art Models in Text Recognition}\label{s:exp-baseline}

For comparison to state-of-the-art methods, we use three models: 
(i) Baek~\etal~\cite{baek2019wrong}
for scene-text recognition;
(ii) Tesseract~\cite{tesseract-ocr},
the industry standard for document OCR;
and (iii)  Chowdhury~\etal~\cite{chowdhury2018efficient}
for handwritten text recognition. 

For (i), we use the open-source models provided, but without the 
transformation module (since documents do not have the scene-text problem of 
non-rectilinear characters). Note, our visual encoder has similar number of 
parameters as in the encoder ResNet of~\cite{baek2019wrong}
(theirs: 6.8M, ours: 4.7M parameters). For (ii) and (iii) we implement the models using the published 
architecture details. Further details of these networks, and the verification 
of our implementations is provided in the \Cref{sec:impl_arch}.

\input{figs/goog1k}
\input{tables/Google_stats}

\subsection{Datasets and Metrics}\label{s:exp-data}
\noindent\textbf{FontSynth.} We take fonts from the MJSynth dataset~\cite{Jaderberg14c} 
and split them into five categories by their appearance attributes 
as determined from their names: (1)~regular, (2)~bold, (3)~italic, 
(4)~light, and (5)~others (\ie, all fonts with none of the first four 
attributes in their name); visualized in \Cref{f:splits+syn} (left).
We use the first four splits to create a training set, and (5) for the test set.
For training, we select 50 fonts at random from each split 
and generate 1000 text-line and glyph images for each font. 
For testing, we use all the 251 fonts in category~(5). 
LRS2 dataset~\cite{Chung16} is used as the text source. 
We call this dataset \emph{FontSynth}; visualization in \Cref{f:splits+syn} 
(right) and further details in the \Cref{sec:impl_arch}.

\noindent\textbf{Omniglot-Seq.}
Omniglot~\cite{lake2015human} consists of 50 alphabets with a total of 
1623 characters, each drawn by 20 different writers.
The original one-shot learning task is defined for \emph{single} characters.
To evaluate our sequence prediction network we generate 
a new \emph{Omniglot-Seq} dataset with \emph{sentence} images as following.
We randomly map alphabets in Omniglot to English, and use them as `fonts'
to render text-line images as in FontSynth above.
We use the original alphabet splits (30 training, 20 test)
and generate data online for training, and 500 lines per alphabet for testing.
\Cref{f:splits+syn} (right) visualizes a few samples. 

\noindent\textbf{Google1000.} Google1000~\cite{google1000} is a standard 
benchmark for document OCR released in ICDAR~2007. 
It constitutes scans of 1000 public domain historical books in English (EN), 
French (FR), Italian (IT) and Spanish (ES) languages; 
\Cref{t:goog1k-stats} provides a summary.
\Cref{f:goog1k} visualizes a few samples from this dataset.
This dataset poses significant challenges due to severe degradation, 
blur, show-through (from behind), inking, fading, oblique text-lines etc. 
Type-faces from $18^{th}$ century are significantly different from modern fonts,
containing old ligatures like \textit{"{\fontfamily{jkpssos}\selectfont st,ct,Qi}"}.
We use this dataset only for evaluation; further details in \Cref{sec:impl_arch}.

\noindent\textbf{Evaluation metrics.}
We measure the character (CER) and word error rates (WER); definitions in
\Cref{sec:metrics}.

\input{tables/fonts_nolm}

\subsection{Overview of Experiments}\label{s:exp-overview}

The goal of our experiments is to evaluate the proposed model against 
state-of-the-art models for text recognition on their generalization ability to 
(1)~novel visual styles \textbf{(VS)} 
(\eg, novel fonts, background, noise \etc), and 
(2)~novel alphabets/languages \textbf{(A)}. 
Specifically, we conduct the following experiments:

\begin{enumerate}
  \item \textbf{VS-1: Impact of number of training fonts.}
    We use FontSynth to study the impact of the number of different training fonts
    on generalization to novel fonts when the exemplars from the testing fonts 
    are provided.
    
  \item \textbf{VS-2: Cross font matching.}
    In this experiment, we do \emph{not} assume access to the testing font.
    Instead of using exemplars from the test font, the most
    similar font from the training set is selected automatically.

  \item \textbf{VS-3: Transfer from synthetic to real data.}
    This evaluates transfer of models trained on synthetic data to real 
    data with historical typeface and degradation. 
    
     \item \textbf{A-1: Transfer to novel Latin alphabets.}
    This evaluates transfer of models trained on English to 
    new Latin languages in Google1000 with additional characters in the alphabet 
    (\eg, French with accented characters).

    \item \textbf{A-2: Transfer to non-Latin glyphs.}
    The above experiments both train and test on Latin alphabets.
    Here we evaluate the generalization of the models trained on English fonts
    to non-Latin scripts in Omniglot-Seq (\eg, from English to Greek).
\end{enumerate}

\subsection{Ablation Study}\label{s:exp-ablation}
We ablate each major component of the proposed model on 
the VS-1 experiment to evaluate its significance.
\Cref{tab:ablation} reports the recognition accuracy on the FontSynth test set 
when trained on one (R) and all four (R+B+L+I) font attributes.
Without the decoder (last row), simply reporting the argmax from the visual 
similarity map reduces to nearest-neighbors or 
one-shot Prototypical~Nets~\cite{snell2017prototypical} method.
This is ineffective for unsegmented text recognition 
(49\% CER vs. 9.4\% CER for the full model).
Excluding the position encoding in the similarity disambiguation 
module leads to a moderate drop. The similarity disambiguation 
\textit{(sim. disamb.)} and linear embedding in class aggregator 
\textit{(agg. embed.)} are both important, especially when the training 
data is limited. With more training data, the advantage brought by these modules 
becomes less significant, while improvement from position encoding 
does not have such a strong correlation with the amount of training data.
\input{tables/ablation}

\subsection{Results}\label{s:exp-results}
\subsubsection{VS-1: Impact of number of training fonts.}\label{s:exp-vs1}
We investigate the impact of the number of training fonts on generalization to 
unseen fonts. For this systematic evaluation, we train the models on an 
increasing number of FontSynth splits--regular, regular + bold, regular + bold +
light, \etc and evaluate on FontSynth test set. These splits correspond to 
increments of 50 new fonts with a different appearance attribute.
\Cref{tab:font_nolm} summarizes the results.
The three baseline SotA models have similar CER when trained on the same amount of data. 
\emph{Tesseract}~\cite{tesseract-ocr} has a slightly better performance but generalizes 
poorly when there is only one attribute in training. Models with an attention-based 
LSTM (Attn. Baek~\etal\cite{baek2019wrong}, Chowdhury~\etal\cite{chowdhury2018efficient}) 
achieve lower WER than those without due to better language modelling.
Notably, our model achieves the same accuracy with 1 training attribute (CER${=}9.4\%$)
as the SotA's with 4 training attributes (CER${>}10\%$), \ie, using 150 (${=}3{\times}50$) less training fonts,
proving the strong generalization ability of the proposed method to unseen fonts.

\textbf{Leveraging visual matching.} Since, our method does not learn class-specific filters 
(unlike conventional discriminatively trained models), but instead is trained for
visual matching, we can leverage non-English glyphs for training. Hence, we further train on Omniglot-Seq data and drastically reduce 
the CER from $5.6\%$ (4 attributes) to  $3.5\%$.  Being able to leverage language-agnostic data for training is a key strength of our model.

\input{graphs/cross-match}

\subsubsection{VS-2: Cross font matching.}\label{s:exp-vs2}
In VS-1 above, our model assumed privileged access to glyphs from the test image.
Here we consider the setting where glyphs exemplars from \emph{training fonts} are used instead.
This we term as \emph{cross matching}, denoted `ours-cross' in \Cref{tab:font_nolm}.
We randomly select 10 fonts from each font attribute and use those as glyph exemplars.
In \Cref{tab:font_nolm} we report the aggregate mean and standard-deviation over all attributes.
To automatically find the best font match, we also measure the similarity 
between the reference and unseen fonts by computing the column-wise entropy in 
the similarity map $\mathcal{S}$ during inference: Similarity scores within each
glyph span are first aggregated to obtain logits 
$\mathcal{P} \in \mathbb{R}^{|\mathcal{A}|\times W'}$, the averaged entropy of 
logits over columns $ \frac{1}{W'}\sum_{i}^{W'} -P_{i}\log(P_i) $ is then used 
as the criterion to choose the best-matched reference font. Performance from the
best-matched exemplar set is reported in `ours-cross selected' in 
\Cref{tab:font_nolm}. With CER close to the last row where test glyphs are 
provided, it is shown that the model does not rely on extra information from the 
new fonts to generalize to different visual styles.
\Cref{f:cross-match} details the performance for each attribute separately.
The accuracy is largely insensitive to particular font attributes---indicating 
the strong ability of our model to match glyph shapes.
Further, the variation decreases as expected as more training attributes are added.

\subsubsection{VS-3: Transfer from synthetic to real data.}\label{s:exp-vs3}
 We evaluate models trained with synthetic data on the real-world Google1000 test set for generalization to novel visual fonts and robustness against degradation
and other nuisance factors in real data.
To prevent giving per test sample specific privileged information to our model, 
we use a common glyph set extracted from Google1000 (visualized in 
\Cref{f:goog1k}). This glyph set is used for \emph{all} test samples, \ie, is 
not sample specific. \Cref{tab:Google_En} compares our model trained on 
FontSynth+Omniglot-Seq against the SotAs. These models trained on modern fonts 
are not able to recognize historical ligatures like long s: \textit{`\fontfamily{jkpkvos}\selectfont s}' and usually classify it as the 
character \textit{`f'}. Further, they show worse ability for handling 
degradation problems like fading and show-through, and thus are outperformed 
by our model, especially when supported by a language model (LM) (CER: ours = $2.4\%$ vs. CTC = $3.14\%$). \\

\input{tables/Google_EN}
\input{figs/new_alpha}

\subsubsection{A-1: Transfer to novel Latin alphabets.}
We evaluate our model trained on English FontSynth + Omniglot-Seq to other 
languages in Google1000, namely, French, Italian and Spanish.
These languages have more characters than English due to accents 
(see~\Cref{t:goog1k-stats}). We expand the glyph set from English to include the
accented glyphs shown in \Cref{f:goog1k}. 
For comparison, we pick the CTC Baek~\etal\cite{baek2019wrong} (the SotA with 
the lowest CER when training data is limited), and adapt it to the new alphabet 
size by fine-tuning the last linear classifier layer on an increasing number of
training samples. \Cref{f:new_alpha} summarizes the results. Images for 
fine-tuning are carefully selected to cover as many new classes as possible. 
For all three languages, at least 5 images with new classes are required in 
fine-tuning to match our performance without fine-tuning; Depending on the 
number of new classes in this language (for French 16 samples are required). 
Note that for our model we do not need fine-tuning at all, just supplying 
exemplars of new glyphs gives a good performance.

\subsubsection{A-2: Transfer to non-Latin glyphs.}
In the above experiments, the models were both trained and tested on
English/Latin script and hence, are not tasked to generalize to completely novel glyph shapes.
Here we evaluate the generalization ability of our model to new glyph shapes by testing the model
trained on FontSynth + Omniglot-Seq on the Omniglot-Seq test set, which consists of novel alphabets/scripts.
We provide our model with glyph exemplars from the randomly generated alphabets (\Cref{s:exp-data}).
Our model achieves CER=$1.8\% / 7.9\%$, WER=$7.6\% / 31.6\%$ (with LM/without 
LM), which demonstrates strong generalization to novel scripts. Note, the 
baseline text recognition models trained on FontSynth (English fonts) cannot 
perform this task, as they cannot process completely new glyph shapes.

%% file: graphs/split.tex
\begin{figure}[t]
	\centering
    \includegraphics[width=\linewidth]{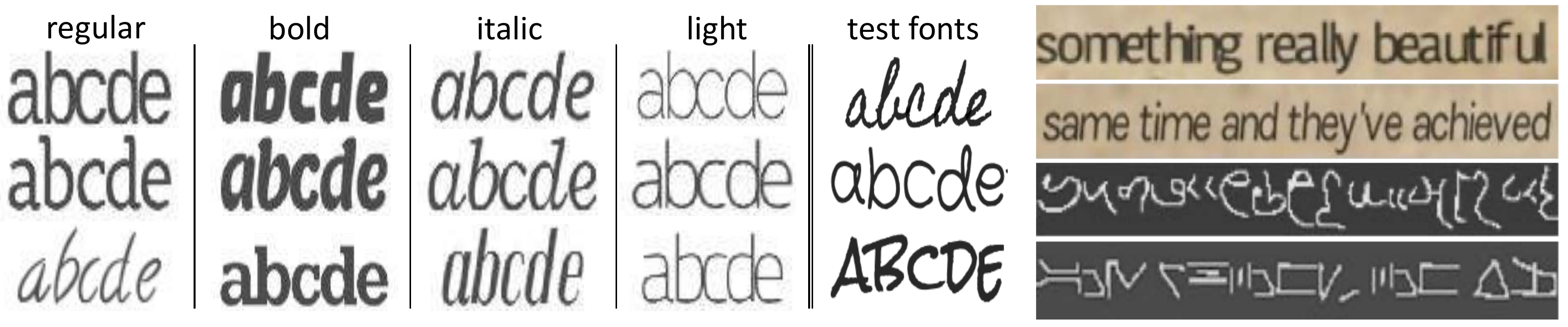}
    \caption{\textbf{Left: FontSynth splits.} Randomly selected fonts from each of the five font categories -- (1)~\emph{regular} (R), (2)~\emph{bold} (B), (3)~\emph{italic} (I), (4)~\emph{light} (L) -- used for generating the synthetic training set, and (5)~\emph{other} (\ie none of the first four) -- used for the test set. \textbf{Right: Synthetic data.} Samples from \emph{FontSynth} \textbf{(top)} generated using fonts from MJSynth~\cite{Jaderberg14c}, and \emph{Omniglot-Seq} \textbf{(bottom)} generated using glyphs from Omniglot~\cite{lake2015human} as fonts (\Cref{s:exp-data}). }\label{f:splits+syn}

\end{figure}

%% file: figs/goog1k.tex
\begin{figure*}[t] 
  \begin{center}
     \includegraphics[width=\linewidth]{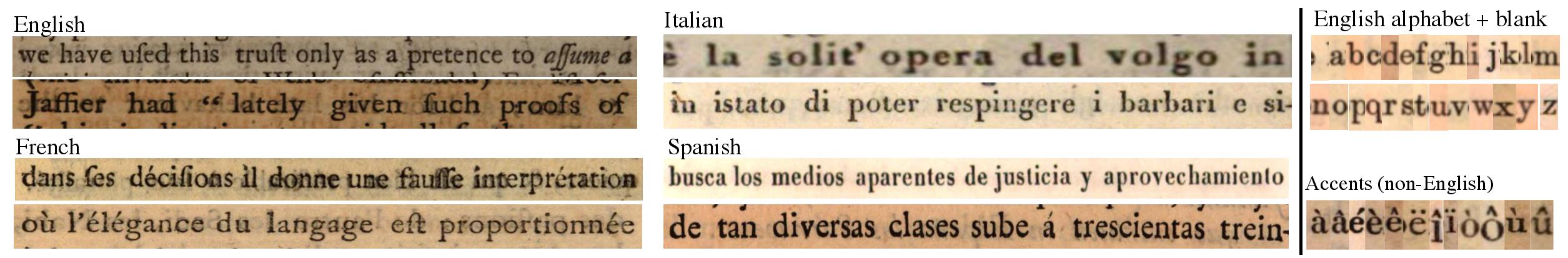}
  \end{center}
   \caption{\textbf{Google1000 printed books dataset.}
   \textbf{(left):} Text-line image samples from the Google1000~\cite{google1000} evaluation set for all the languages, namely, English, French, Italian and Spanish. \textbf{(right):} \emph{Common} set of glyph exemplars used in our method for \emph{all} books in the evaluation set for English and accents for the other languages.}
  \label{f:goog1k}
\end{figure*}

%% file: tables/Google_stats.tex
\setlength\intextsep{6pt}
\begin{table}

\centering
 \resizebox{0.6\linewidth}{!}{%
\begin{tabular}{@{}lrlrrr@{}}
\toprule
\multicolumn{1}{c}{language $\rightarrow$} & \multicolumn{2}{c}{EN}  & \multicolumn{1}{c}{FR} & \multicolumn{1}{c}{IT} & \multicolumn{1}{c}{ES} \\ \midrule
\# books                                   & \multicolumn{2}{r}{780} & 40                     & 40                     & 140                             \\
alphabet size                              & \multicolumn{2}{r}{26}  & 35                     & 29                     & 32                              \\
\% accented letters                        & \multicolumn{2}{r}{0}   & 2.6                    & 0.7                    & 1.5                             \\ \bottomrule
\end{tabular}%
}
\caption{\textbf{Google1000 dataset summary.} 
	Total number of books, alphabet size and percentage of letters with accent 
	(counting accented characters a new) for various languages in 
	the Google1000.}\label{t:goog1k-stats}
\vspace{-5mm}
\end{table}

%% file: tables/fonts_nolm.tex
\begin{table*}[tbp]	
  \setlength{\tabcolsep}{8pt}
  \renewcommand{\arraystretch}{0.6}
	\centering
	\resizebox{\textwidth}{!}{%
	\begin{tabular}{crccc|cc|cc|cc|cc}
		\toprule
		\multicolumn{2}{c}{} & \multicolumn{1}{c|}{} &  &  &  &  &  &  &  &  &  &  \\
		\multicolumn{2}{l}{training set $\rightarrow$} & \multicolumn{1}{c|}{} & \multicolumn{2}{c|}{\textbf{{\small R}}} & \multicolumn{2}{c|}{\textbf{{\small R+B}}} & \multicolumn{2}{c|}{\textbf{{\small R+B+L}}} & \multicolumn{2}{c|}{\textbf{{\small R+B+L+I}}} & \multicolumn{2}{c}{\textbf{{\small R+B+L+I+OS}}} \\ 
		\multicolumn{2}{c}{} & \multicolumn{1}{c|}{} &  &  &  &  &  &  &  &  &  &  \\\hline
		\multicolumn{2}{l}{model} & \multicolumn{1}{c|}{\scriptsize{\textbf{\begin{tabular}[c]{@{}c@{}}	\rule{0pt}{3ex} test \\ glyphs\\ known\end{tabular}}}} & \textbf{\small CER} & \textbf{\small WER} & \textbf{\small CER} & \textbf{\small WER} & \textbf{\small CER} & \textbf{\small WER} & \textbf{\small CER} & \textbf{\small WER} & \textbf{\small CER} & \textbf{\small WER} \\ \hline
		\multicolumn{2}{l}{\begin{tabular}[l]{@{}c@{}}\rule{0pt}{3ex}CTC Baek~\etal~\cite{baek2019wrong}\end{tabular}} & \multicolumn{1}{c|}{\xmark} & 17.5 & 46.1 & 11.5 & 30.3 & 10.4 & 28.2 & 10.4 & 27.7 & --- & --- \\
		\multicolumn{2}{l}{\begin{tabular}[c]{@{}c@{}}Attn. Baek~\etal~\cite{baek2019wrong}\end{tabular}} & \multicolumn{1}{c|}{\xmark} & 16.5 & 41.0 & 12.7 & 34.5 & 11.1 & 27.4 & 10.3 & 23.6 & --- & --- \\
		\multicolumn{2}{l}{\multirow{2}{*}{Tesseract~\cite{tesseract-ocr}}} & \multicolumn{1}{c|}{\multirow{2}{*}{\xmark}} & \multirow{2}{*}{19.2} & \multirow{2}{*}{48.6} & \multirow{2}{*}{12.3} & \multirow{2}{*}{37.0} & \multirow{2}{*}{10.8} & \multirow{2}{*}{31.7} & \multirow{2}{*}{9.1} & \multirow{2}{*}{27.8} & \multirow{2}{*}{---} & \multirow{2}{*}{---} \\
		\multicolumn{2}{c}{} & \multicolumn{1}{c|}{} &  &  &  &  &  &  &  &  &  &  \\
		\multicolumn{2}{l}{{Chowdhury~\etal~\cite{chowdhury2018efficient}}} & \multicolumn{1}{c|}{{\xmark}} & {16.2} & {39.1} & {12.6} & {28.6} & {11.5} & {29.5} & {10.5} & {24.2} & {---} & {---} \\
		\midrule
		\multicolumn{1}{l}{\multirow{2}{*}{\textbf{ours-cross}}} & mean & \multicolumn{1}{c|}{\multirow{2}{*}{\xmark}} & 11.0 & 33.7 & 9.3 & 30.8 & 9.1 & 28.6 & 7.6 & 22.2 & 7.0 & 25.8 \\
		& (std) & \multicolumn{1}{c|}{} & \small{(2.9)} &  \small{(9.8)} &  \small{(1.4)} &  \small{(5.9)} &  \small{(1.1)} &  \small{(2.2)} &  \small{(0.2)} &  \small{(0.9)} &  \small{(0.9)} &  \small{(3.7)} \\
		\multicolumn{2}{c}{} & \multicolumn{1}{c|}{} &  &  &  &  &  &  &  &  &  &  \\
		\multicolumn{1}{l}{\textbf{ours-cross}} & selected & \multicolumn{1}{c|}{\xmark} & {9.8} & {30.0} & {8.4}& {29.4} & {8.4} & {27.8} & {7.2} & {21.8} & {5.3} & {18.3}  \\
		\multicolumn{2}{l}{{\textbf{ours}}} & \multicolumn{1}{c|}{{\cmark}} & {\textbf{9.4}} & {\textbf{30.2}} & {\textbf{8.3}} & {\textbf{28.8}} & {\textbf{8.1}} & {\textbf{27.3}} & {\textbf{5.6}} & {\textbf{22.4}} & {\textbf{3.5}} & {\textbf{12.8}}  \\
		\bottomrule
	\end{tabular}}
	\caption{\textbf{VS-1, VS-2: Generalization to novel fonts with/without known test glyphs and increasing number of training fonts.} 
	The error rates (in $\%$; $\downarrow$ is better) on \textbf{FontSynth} test set. %
	\emph{Ours-cross} stands for cross font matching where test glyphs are unknown and training fonts are used as glyph exemplars, mean and standard-dev reported when the exemplar fonts are randomly chosen from the training set, while \emph{selected} shows results from the best matched exemplars automatically chosen based on confidence measure.
	\emph{R}, \emph{B}, \emph{L} and \emph{I} correspond to the FontSynth training splits;
	\emph{OS} stands for the Omniglot-Seq dataset (\Cref{s:exp-data}).}\label{tab:font_nolm}
\end{table*}

%% file: tables/ablation.tex
\setlength\intextsep{10pt}
\begin{table}[b]
    \setlength{\tabcolsep}{3pt}

	\centering
	\resizebox{1.0\linewidth}{!}{%
	\begin{tabular}{cccc|cccc}
	\hline
	\multicolumn{1}{c|}{\multirow{3}{*}{\begin{tabular}[c]{@{}c@{}}sim. enc.\\ $\mathcal{S}$\end{tabular}}} & \multicolumn{2}{c|}{\multirow{2}{*}{sim. disamb.}} & \multirow{3}{*}{\begin{tabular}[c]{@{}c@{}}agg. \\ embed.\end{tabular}} & \multicolumn{4}{c}{training  data} \\ \cline{5-8} 
	\multicolumn{1}{c|}{} & \multicolumn{2}{c|}{} &  & \multicolumn{2}{c|}{R} & \multicolumn{2}{c}{R+B+L+I} \\ \cline{2-3}
	\multicolumn{1}{c|}{} & \multicolumn{1}{c|}{pos. enc.} & \multicolumn{1}{c|}{self-attn} &  & CER & \multicolumn{1}{c|}{WER} & CER & WER \\ \hline
	\cmark & \cmark & \cmark & \cmark & \textbf{9.4} & \multicolumn{1}{c|}{\textbf{30.1}} & \textbf{5.6} & \textbf{22.3} \\ \hline
	\cmark & \xmark & \cmark & \cmark & 11.8 & \multicolumn{1}{c|}{37.9} & 7.9 & 22.9 \\
	\cmark & \xmark & \xmark & \cmark & 23.9 & \multicolumn{1}{c|}{68.8} & 13.0 & 52.0 \\
	\cmark & \cmark & \cmark & \xmark & 22.9 & \multicolumn{1}{c|}{65.8} & 8.5 & 26.4 \\
	\cmark & \xmark & \xmark & \xmark & 25.8 & \multicolumn{1}{c|}{63.1} & 18.4 & 45.0 \\
	\cmark & --- & --- & --- & 49.0 & \multicolumn{1}{c|}{96.2} & 38.3 & 78.9 \\ \hline
\end{tabular}%
 }
	\caption{\textbf{Model component analysis.} 
	The first row corresponds to the full model; the
	last row corresponds to reading out characters using the CTC decoder from the output of the visual encoder. \emph{R}, \emph{B}, \emph{L} and \emph{I} correspond to the FontSynth training
	splits: Regular, Bold, Light and Italic respectively.}
\label{tab:ablation}
\end{table}

%% file: graphs/cross-match.tex
\setlength\intextsep{3pt}
\begin{figure}
	\centering
    \includegraphics[width=0.88\linewidth]{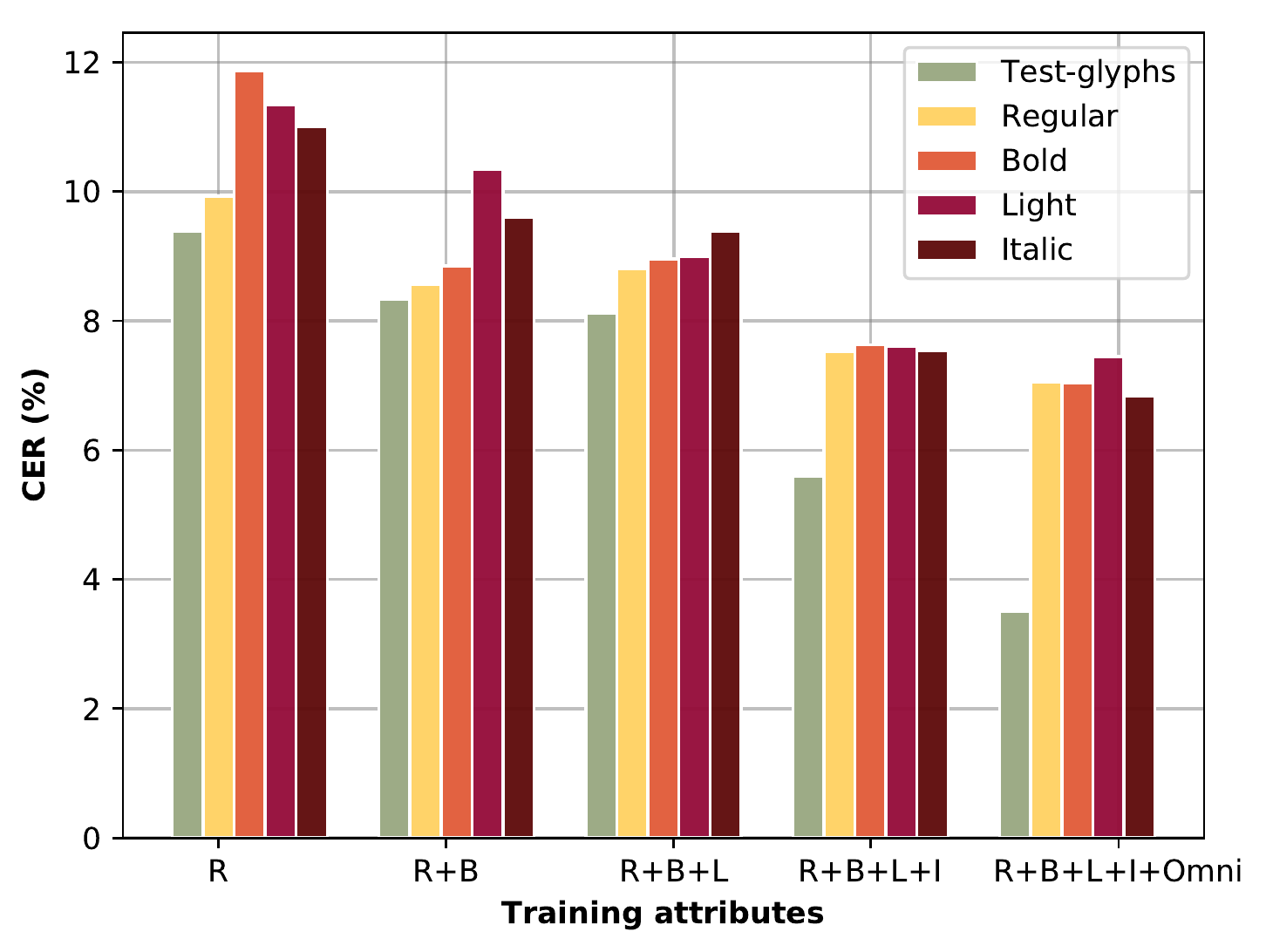}
    \caption{\textbf{VS-2: Performance on FontSynth when using different exemplars for cross-font matching.} When using glyphs from other fonts (\eg regular, bold \etc.) as exemplars, the accuracy is insensitive to the attribute. Further, using test-glyphs as exemplars significantly improve the performance.
    On the $x$-axis we show the FontSynth training splits (\Cref{f:splits+syn} left).}\label{f:cross-match}  
\end{figure}
\vspace{-3mm}

%% file: tables/Google_EN.tex
\begin{table}
  \setlength{\tabcolsep}{3pt}
	\centering
	\resizebox{\linewidth}{!}{%
	\begin{tabular}{@{}lcccccccccc@{}}
	\toprule
	\multicolumn{1}{c}{} & \multicolumn{2}{c|}{\begin{tabular}[c|]{@{}c@{}}{\small CTC} \\ {\small Baek~\cite{baek2019wrong}}\end{tabular}} & \multicolumn{2}{c|}{\begin{tabular}[c]{@{}c@{}}{\small Attn.} \\ {\small Baek~\cite{baek2019wrong}}\end{tabular}} & \multicolumn{2}{c|}{\begin{tabular}[c]{@{}c@{}}{\small Tesseract} \\ {\small \cite{tesseract-ocr}}\end{tabular}}& \multicolumn{2}{c|}{\begin{tabular}[c]{@{}c@{}}{\small Ch.~\etal}\\{\small \cite{chowdhury2018efficient}}\end{tabular}}& \multicolumn{2}{c}{\textbf{ours}} \\ \midrule
	{\small LM}          &  \multicolumn{1}{c}{\xmark}       & \multicolumn{1}{c|}{\cmark} &  \multicolumn{1}{c}{\xmark} & \multicolumn{1}{c|}{\cmark}  &  \multicolumn{1}{c}{\xmark} & \multicolumn{1}{c|}{\cmark} & \multicolumn{1}{c}{\xmark} & \multicolumn{1}{c|}{\cmark}    &  \multicolumn{1}{c}{\xmark }         & \multicolumn{1}{c}{\cmark}          \\ \midrule
	{\small CER}         &   \multicolumn{1}{c}{3.5}           & \multicolumn{1}{c|}{3.1}   &  \multicolumn{1}{c}{5.4}   & \multicolumn{1}{c|}{5.4}     &  \multicolumn{1}{c}{4.7}   & \multicolumn{1}{c|}{3.8}   &  \multicolumn{1}{c}{ 5.5}    & \multicolumn{1}{c|}{5.6}   & \textbf{3.1}    & \multicolumn{1}{c}{ \textbf{2.4}} \\[2mm]
	{\small WER}         & \textbf{12.9} & \multicolumn{1}{c|}{11.4}   &    \multicolumn{1}{c}{13.1}   & \multicolumn{1}{c|}{13.8}    &  \multicolumn{1}{c}{15.9}   & \multicolumn{1}{c|}{12.2} &  \multicolumn{1}{c}{14.9}  & \multicolumn{1}{c|}{15.6}  & \multicolumn{1}{c}{14.9}            & \multicolumn{1}{c}{\textbf{8.0}}   \\ \bottomrule
	\end{tabular}}
	\caption{\textbf{VS-3: Generalization from synthetic to real data.} Mean error rates (in $\%$; $\downarrow$ is better) on Google1000 English document for models trained only on synthetic data (\Cref{s:exp-vs3}).
	LM stands for 6-gram language model.}\label{tab:Google_En}
\end{table}
\vspace{-8mm}

%% file: figs/new_alpha.tex
\begin{figure}[b]
	\centering
    \includegraphics[width=1.0\linewidth]{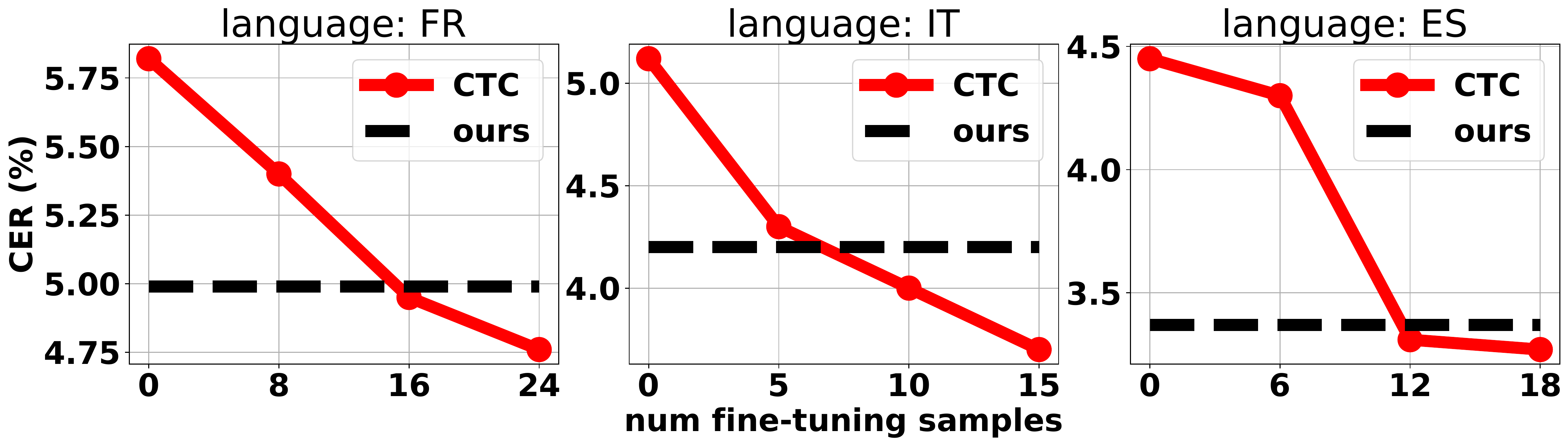}
    \caption{\textbf{A-2: Transfer to novel alphabets in Google1000.} 
    We evaluate models trained over the English alphabet on novel languages
    in the Google1000 dataset, namely, French, Italian and Spanish.
    CER is reported (in $\%$; $\downarrow$ is better).}  
    \label{f:new_alpha}
\end{figure}

%% file: sections/conclusion.tex
\section{Conclusion} \label{s:conc}
We have developed a method for text recognition which generalizes to novel 
visual styles (\eg, fonts, colors, backgrounds \etc), and is not tied to a 
particular alphabet size/language. It achieves this by recasting the classic 
text recognition as one of visual matching, and we have demonstrated that the 
matching can leverage random shapes/glyphs (\eg, Omniglot) for training. Our 
model is perhaps the first to demonstrate one-shot sequence recognition, and 
achieves superior generalization ability as compared to conventional text 
recognition methods without requiring expensive adaptation/fine-tuning.
Although the method has been demonstrated for text recognition, it is applicable
to other sequence recognition problems like speech and action recognition.

\noindent\textbf{Acknowledgements.} This research is funded by a Google-DeepMind Graduate Scholarship and the EPSRC Programme Grant Seebibyte EP/M013774/1. We would like to thank Triantafyllos Afouras, Weidi Xie, Yang Liu and Erika Lu for discussions and proof-reading.

%% file: supp/supp.tex
\input{supp/include}

\input{supp/sections/contents}

\vspace{10mm}

\input{supp/sections/metrics}

\vspace{10mm}
\input{supp/sections/ablation}
\clearpage

\input{supp/sections/visualizations}

\clearpage

\input{supp/sections/arch}
\pagebreak

\input{supp/sections/scene}
\pagebreak
\input{supp/sections/dataset.tex}

\clearpage

%% file: supp/include.tex
\newcolumntype{L}{>{\raggedright\arraybackslash}X}
\newcolumntype{R}{>{\raggedleft\arraybackslash}X}
\newcolumntype{Y}{>{\centering\arraybackslash}X}

\newcommand{\hboxSpace}[1]{
    {\setlength{\fboxrule}{0pt}%
     \fbox{\hspace{#1}}}%
}

\newlength\swidth
\newlength\lheight
\newlength\gwidth
\newlength\twidth
\newlength\sswidth

\setlength\swidth{0.3\textwidth}
\setlength\sswidth{0.15\textwidth}
\setlength\lheight{4mm}
\setlength\gwidth{0.5\textwidth-1.5\swidth-0.5\lheight}
\setlength\twidth{3\swidth+3mm}

\newcommand{\setSimWidth}[1]{
    \setlength\swidth{#1\textwidth}
    \setlength\gwidth{0.5\textwidth-1.5\swidth-0.5\lheight}
    \setlength\twidth{3\swidth+3mm}
}

\newcommand{\simfig}[3]{
\begin{center}
		\rotatebox[origin=rB]{270}{\includegraphics[height=\lheight,,width=\swidth]{supp/media/#1_glyph}}
		\includegraphics[height=\swidth,width=\swidth]{supp/media/#1_S0}
		\includegraphics[height=\swidth,width=\swidth]{supp/media/#1_S1}
		\includegraphics[height=\swidth,width=\swidth]{supp/media/#1_S2}\\
		\hspace{5mm}\includegraphics[height=\lheight,width=\swidth]{supp/media/#1_line}
		\includegraphics[height=\lheight,width=\swidth]{supp/media/#1_line}
		\includegraphics[height=\lheight,width=\swidth]{supp/media/#1_line}

		\hboxSpace{\gwidth}%
		\begin{tabularx}{\twidth}{LL}
			\centering
			\scriptsize{\textbf{GT:} #2 }&	\scriptsize{\textbf{Pred:} #3}
		\end{tabularx}
\end{center}}

\newcommand{\simheaderfig}[3]{
        \hboxSpace{\gwidth}%
		\begin{tabularx}{\twidth}{YYY}
			\scriptsize{\textbf{$\mathcal{S}$ (from visual encoder $\Phi$)}} & \scriptsize{\textbf{$\mathcal{S}$ + position encoding} }& \scriptsize{\textbf{enhanced $\mathcal{S}^*$ (after self-attention)}}
		\end{tabularx}
	  \simfig{#1}{#2}{#3}
}

\newcommand{\scenefig}[7]{
	\begin{center}
		\rotatebox[origin=rB]{270}{\includegraphics[height=\lheight,width=\swidth]{supp/media/#1}}
		\includegraphics[height=\swidth,width=\sswidth]{supp/media/#2_sim}	\hspace{\sswidth}
		\rotatebox[origin=rB]{270}{\includegraphics[height=\lheight,width=\swidth]{supp/media/#1}}
		\includegraphics[height=\swidth,width=\sswidth]{supp/media/#3_sim}	\hspace{\sswidth}
		\rotatebox[origin=rB]{270}{\includegraphics[height=\lheight,width=\swidth]{supp/media/#1}}
		\includegraphics[height=\swidth,width=\sswidth]{supp/media/#4_sim}	\hspace{\sswidth} \\
			\hspace{2mm}	\includegraphics[height=\lheight,width=\sswidth]{supp/media/#2_line} \hspace{\sswidth}
			\hspace{5mm}\includegraphics[height=\lheight,width=\sswidth]{supp/media/#3_line}\hspace{\sswidth}
		  	\hspace{6mm}\includegraphics[height=\lheight,width=\sswidth]{supp/media/#4_line}\hspace{\sswidth}
	\vspace{-0.5mm}
	\resizebox{1.15\linewidth}{!}{%
	\setlength{\tabcolsep}{39pt} %
			\begin{tabularx}{\linewidth}{LLL}

				\multicolumn{1}{c}{	\scriptsize{\textbf{GT:} #5 }  \scriptsize{\textbf{Pred:} #5}} & \multicolumn{1}{c}{	\scriptsize{\textbf{GT:} #6 } 	\scriptsize{\textbf{Pred:} #6}} & \multicolumn{1}{c}{	\scriptsize{\textbf{GT:} #7 }	\scriptsize{\textbf{Pred:} #7}}
		
		\end{tabularx}
}
\end{center}}

\newcommand{\wrongscenefig}[9]{
	\begin{center}
		\rotatebox[origin=rB]{270}{\includegraphics[height=\lheight,width=\swidth]{supp/media/#1}}
		\includegraphics[height=\swidth,width=\sswidth]{supp/media/#2_sim}	\hspace{\sswidth}
		\rotatebox[origin=rB]{270}{\includegraphics[height=\lheight,width=\swidth]{supp/media/#1}}
		\includegraphics[height=\swidth,width=\sswidth]{supp/media/#3_sim}	\hspace{\sswidth}
		\rotatebox[origin=rB]{270}{\includegraphics[height=\lheight,width=\swidth]{supp/media/#1}}
		\includegraphics[height=\swidth,width=\sswidth]{supp/media/#4_sim}	\hspace{\sswidth} \\
		\hspace{2mm}	\includegraphics[height=\lheight,width=\sswidth]{supp/media/#2_line} \hspace{\sswidth}
		\hspace{5mm}\includegraphics[height=\lheight,width=\sswidth]{supp/media/#3_line}\hspace{\sswidth}
		\hspace{6mm}\includegraphics[height=\lheight,width=\sswidth]{supp/media/#4_line}\hspace{\sswidth}
		\vspace{-0.5mm}
		\resizebox{1.1\linewidth}{!}{%
				\setlength{\tabcolsep}{39pt} %
			\begin{tabularx}{\linewidth}{LLL}
				\multicolumn{1}{c}{	\scriptsize{\textbf{GT:} #5 }  \scriptsize{\textbf{Pred:} #5}} & \multicolumn{1}{c}{	\scriptsize{\textbf{GT:} #6 } \scriptsize{\textbf{Pred:} \textcolor{red}{#7}}} & \multicolumn{1}{c}{	\scriptsize{\textbf{GT:} #8 }	\scriptsize{\textbf{Pred:} \textcolor{red}{#9}}}
				
			\end{tabularx}
		}
\end{center}}

%% file: supp/sections/contents.tex
\section*{Contents}\label{sec:overview}

\begin{tabularx}{\textwidth}{Ll}
 Evaluation Metrics            		 				& \Cref{sec:metrics}  \\
  Ablation Study                                & \Cref{sec:ablation}  \\
  Examples and Visualizations 					   & \Cref{sec:visual}    \\
  Implementation of SotA Models					   & \Cref{sec:impl_arch}      \\
  Performance on Scene Text 							& \Cref{sec:scene}  \\
  Dataset Details                               & \Cref{sec:dataset}  \\
\end{tabularx}

%% file: supp/sections/metrics.tex
\section{Evaluation Metrics}\label{sec:metrics}
We measure the character (CER) and word error rates (WER):

\begin{equation*}
    \text{CER} = \frac{1}{N}\sum\limits_{i=1}^N \frac{\operatorname{EditDist}\left(\by_{\text{gt}}^{(i)}, \by_{\text{pred}}^{(i)}\right)}{\operatorname{Length}\left(\by_{\text{gt}}^{(i)}\right)}
\end{equation*}
  where, $\by_{\text{gt}}^{(i)}$ and $\by_{\text{pred}}^{(i)}$ are the $i^{\text{th}}$ ground-truth and predicted
  strings respectively in a dataset containing $N$ strings;
  $\operatorname{EditDist}$ is the \emph{Levenshtein distance}~\cite{Levenshtein66};
  $\operatorname{Length}\left(\by_{\text{gt}}\right)$ is the number of characters in $\by_{\text{gt}}$.
WER is computed as CER above with words (\ie contiguous characters separated by whitespace) in place of characters as tokens.

%% file: supp/sections/ablation.tex
\section{Ablation study}\label{sec:ablation}

\subsection{Ablation on modules at a larger scale}
In \Cref{s:exp-ablation}, we ablated each major component of the proposed model architecture
to evaluate its relative contribution to the model's performance.
However, there the training set was limited to only one FontSynth attribute (regular fonts).
Here in \Cref{tab:supp_ablation}, we ablate the same model components, but couple it with increasing number
of training fonts from the FontSynth dataset (\Cref{s:exp-data}), still evaluating on the FontSynth test set.

\input{supp/tables/ablation}

Predictions straight from the output of the visual encoder (not using the decoder) are quite noisy (row 6). Using a class aggregator (`agg. embed') lends robustness to noise in the similarity maps which leads to consistent improvements across all training data settings (compare rows 3 and 5). 
Using position encoding (`pos. enc.') which encodes glyph extents also leads to consistent improvements (compare rows 1 and 2).
Self-attention in the similarity disambiguator is a critical component, without which error increases at least twofold (compare rows 2 and 3). 
With increasing training data, the performance generally improves for all the ablated versions of the model explored here. However, it is evident that all the model components retain their functional significance even with large amounts of training data.

\subsection{Ablation on balance of losses}\label{s:loss_ablation}
\Cref{f:weight} shows the impact of $\lambda$ when training with one FontSynth attribute, where $\lambda$ is the weight on $L_{sim}$ as in \Cref{e:loss_tot}. The use of  $L_{sim}$ is essential as the model does not converge when its ratio is smaller than 0.05, labelled by `x' mark in \Cref{f:weight}. The CER is also increased by about 1\% when its weight is five times larger than that on  $L_{pred}$. 
\input{supp/figs/weight_ablation.tex}

%% file: supp/tables/ablation.tex
\setlength{\tabcolsep}{2pt} %
\begin{table}[h]
\centering
\begin{tabular}{@{}ccccc|cc|cc|cc|cc|cc}
    \toprule
 &	\multicolumn{1}{c|}{\multirow{3}{*}{\begin{tabular}[c]{@{}c@{}}sim. enc.\\ $\mathcal{S}$\end{tabular}}} & \multicolumn{2}{c|}{\multirow{2}{*}{sim. disamb.}} & \multirow{3}{*}{\begin{tabular}[c]{@{}c@{}}agg. \\ embed.\end{tabular}} && \multicolumn{9}{c}{training  data} \\ \cline{6-15} 
	&\multicolumn{1}{c|}{} & \multicolumn{2}{c|}{} &  &  \multicolumn{2}{c|}{\small \textbf{R}} & \multicolumn{2}{c|}{\small \textbf{R+B}}& \multicolumn{2}{c|}{\small \textbf{R+B+L}}& \multicolumn{2}{c|}{\small \textbf{R+B+L+I}}& \multicolumn{2}{c}{\small \textbf{R+B+L+I+OS}}\\  \cline{3-4}
  &&\multicolumn{1}{|c|}{pos. enc.}& \multicolumn{1}{c|}{self-attn} & &  {\small\textbf{CER}} & {\small\textbf{WER}}  &  {\small\textbf{CER}} & {\small\textbf{WER}}&  {\small\textbf{CER}} & {\small\textbf{WER}}&  {\small\textbf{CER}} & {\small\textbf{WER}}&  {\small\textbf{CER}} & {\small\textbf{WER}} \\ \midrule
  1. & \cmark & \cmark   & \cmark & \cmark & \textbf{9.4} & \textbf{30.2} & \textbf{8.3} & \textbf{28.8}& \textbf{8.1} & \textbf{27.3} & \textbf{5.6} & \textbf{22.3}   & \textbf{3.5} & \textbf{12.8}\\ \midrule
  2. & \cmark & \xmark   & \cmark & \cmark & 11.8  & 37.9  & 12.4 &41.1 &9.4 &27.6 &7.9 &22.9 &4.2 &16.8\\
  3. & \cmark  & \xmark  & \xmark  & \cmark & 23.9  &  68.8  & 15.4 & 39.4 &13.2 & 43.5& 13.0& 52.0&11.4 &49.6\\
    4. & \cmark  & \cmark  & \cmark & \xmark & 22.9  & 65.8  & 8.6 & 34.3 & 8.3 &28.5 &8.5 &26.4 & 4.5 & 20.2 \\
  5. & \cmark  & \xmark  & \xmark  & \xmark  & 25.8  & 63.1   & 19.3 & 54.9 & 18.5 & 48.7& 18.4& 45.0& 20.1&62.1\\
  6. & \cmark  & ---  & ---   & ---  & 49.0   & 96.2 & 33.7& 67.7& 31.8& 72.0& 38.3 & 78.9 & 41.8 & 98.0 \\ \bottomrule
  \end{tabular}%
\caption{\textbf{Model component analysis.} We ablate various components of the model and report performance
on the FontSynth test set when trained on an increasing number of FontSynth attributes. The first row corresponds to the full model; the
last row (\#6) corresponds to reading out characters using the CTC decoder from the output of the visual encoder. `R', `B', `L' and `I' correspond to the FontSynth training splits: `Regular', `Bold', `Light' and `Italic'; while `OS' stands for the Omniglot-Sequence dataset. }
\label{tab:supp_ablation}
\end{table}

%% file: supp/figs/weight_ablation.tex
\begin{figure}[h]
	\centering
    \includegraphics[width=0.5\linewidth]{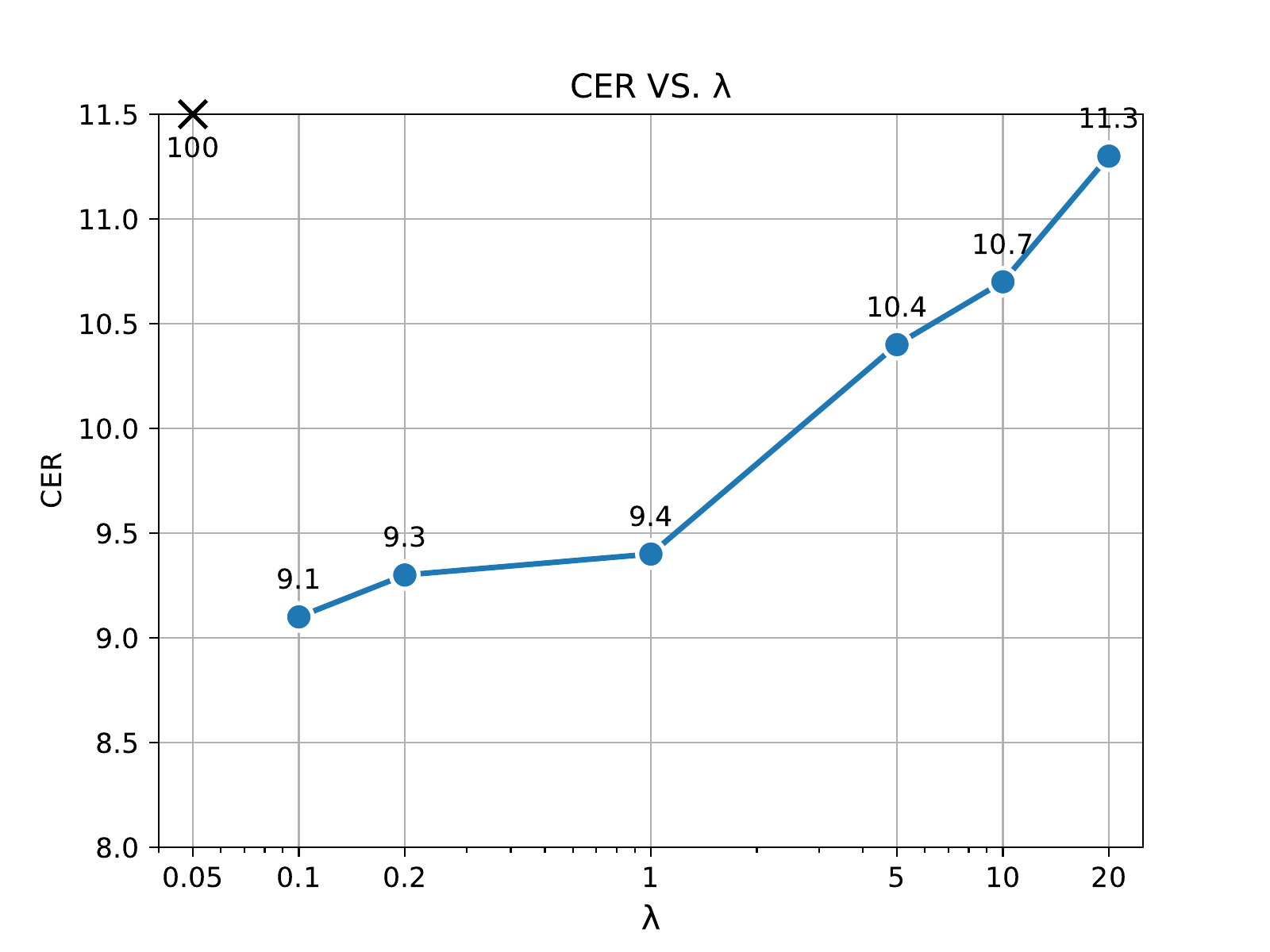}
    \caption{\textbf{Impact of $\lambda$  on CER when training with one FontSynth attribute.} }
  \label{f:weight}
\end{figure}

%% file: supp/sections/visualizations.tex
\section{Examples and Visualizations}\label{sec:visual}
In \Cref{f:vs1,f:vs2,f:vs3,f:vs4,f:a1} we show the similarity maps at various stages in the model
for all the main experiments presented in the paper \textbf{VS1-3, A1-2} (\Cref{s:exp-overview}).
The three similarity maps correspond to (from left to right): 
(1)~$\mathcal{S}$ similarity map from cosine similarity of features from the visual encoder ($\Phi$), 
(2)~similarity map after local position encoding, 
and (3)~$\mathcal{S}^*$ similarity map after the self-attention module. 
All maps are in the range [-1,1].

By comparing maps at different stages, it is clear that the \emph{position encoding}
removes some ambiguities in shape matching, \eg `w'  and `v', `m' and `n', 
by using the positional and width information of exemplars. 
Further, \emph{self-attention} is crucial as it compares the confidence across
all glyphs and suppresses the lower confidence glyphs, while boosting
the higher confidence glyphs, as is evident by increased contrast of the maps.

\Cref{f:detail} shows a zoomed in version of $\mathcal{S}^*$. 
The self-attention module (coupled with the class aggregator; see fig.~2 in the paper)
introduces the boundary token for the CTC loss~\cite{graves2006connectionist} to separate characters, especially repeated characters.
Training with CTC is known to result in peaky distribution along the sequence -- it predicts character classes when the confidence is very high, usually at the center of the character, while predicting all the other positions between characters as boundary token.  This effect can be seen from the top rows of the map, where white pixels correspond to boundary token and the gaps are where the model looks at the central column in each `glyph-square' with high confidence.

\vfill
\input{supp/figs/detailed.tex}

\vfill
\clearpage

\input{supp/figs/VS1}
 \clearpage

\input{supp/figs/VS2}
\clearpage

\input{supp/figs/VS3}
\clearpage

\input{supp/figs/A1}
\clearpage

\input{supp/figs/VS4}
\clearpage

%% file: supp/figs/detailed.tex
\newlength\dwidth
\setlength\dwidth{0.7\linewidth}

\begin{figure}[h]
  \begin{center}
   \rotatebox[origin=rB]{270}{\includegraphics[width=0.7\dwidth]{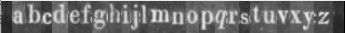}}
   \includegraphics[height=0.7\dwidth,width=0.7\dwidth]{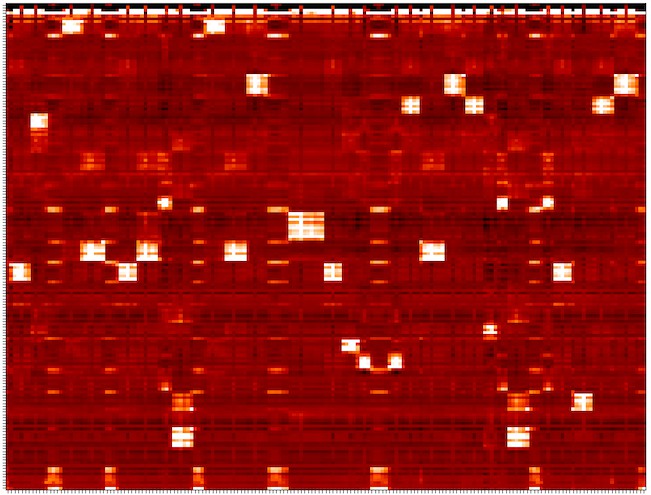}
   \includegraphics[height=0.7\dwidth]{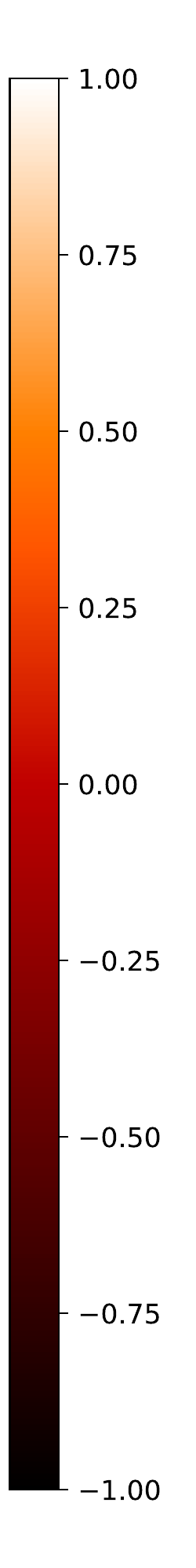}\\
   \hspace{-2mm}\includegraphics[width=0.69\dwidth]{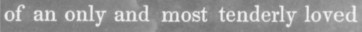}

  \end{center}
     \caption{\textbf{An example of a similarity map $\mathcal{S}^*$ after the self-attention module.}}
  \label{f:detail}
\end{figure}

%% file: supp/figs/VS1.tex
\begin{figure}[h]
\centering
     \caption{\textbf{Experiment VS-1: Generalization to novel fonts with known glyph exemplars.} 
     We use FontSynth to study generalization to novel fonts when the exemplars from test fonts are provided in matching. Challenging samples from the FontSynth test set (novel fonts) 
     are visualized below. Note the marked improvement in the confidence and coherence of the similarity maps through the through various processing stages.}\label{f:vs1}

  \vspace{6mm}
         \simheaderfig{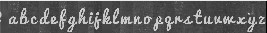}{purpose for coming back}{purpose for coming back}
    \vfill
    
    \simfig{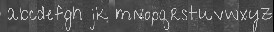}{known each other for years}{known each other for years}
    \vfill
    
    \simfig{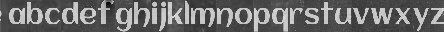}{a great script and i think it was}{a great script and i think it was}
    \vfill

\end{figure}

%% file: supp/figs/VS2.tex
\begin{figure}[!h]
\centering
    \caption{\textbf{Experiment VS-2: Cross font matching.} Text recognition on the FontSynth test set with \emph{cross-font matching} -- using exemplars from the training set when we do not have access to the test font. The model succeeds at cross-font matching even when the exemplars and line images are in very different visual styles. A difficulty occurs when the same glyph in the two fonts are not quite visually similar, \eg due to differences in upper vs.\ lower case. One example is shown in the last row where the capital `Q' is matched to lower case `o' instead of `q'.
    However, using consensus from multiple training exemplars can help overcome this limitation.}\label{f:vs2}
  \vspace{6mm}
  \begin{center}
     \simheaderfig{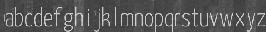}{recommend in my bradshaw}{recommend in my bradshaw}
\vfill

\simfig{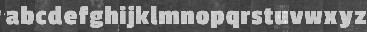}{beaches of normandy}{beaches of normandy}
\vfill

\simfig{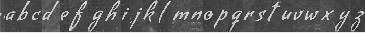}{cosmetically improved quite}{cosmetically improved ouite}
\vfill
  \end{center}
\end{figure}

%% file: supp/figs/VS3.tex
\begin{figure}[!h]
    \centering
     \caption{\textbf{Experiment VS-3: Transfer from synthetic to real data.} We test our model trained purely on \emph{synthetic data} for generalization to \emph{real world} Google1000 English books.
     The model is robust to nuisance factors prevalent in scans of real historical books, \eg over-exposure and low contrast, as well as degradation and show-through.}\label{f:vs3}
     \vspace{6mm}
     \simheaderfig{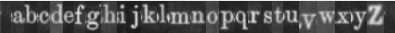}{of styled the protector}{of styled the protector}
\vfill

\simfig{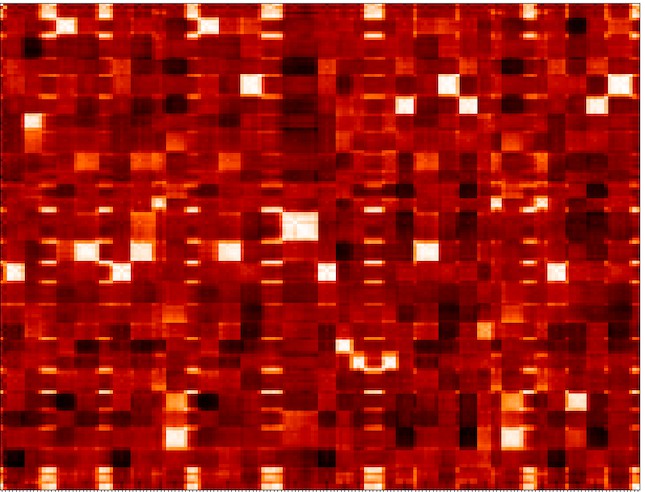}{of an only and most tenderly loved}{of an only and most tenderly loved}
\vfill

\simfig{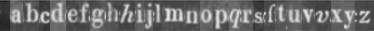}{yode to seek his reverend}{yode to seek his reverend}
\vfill
     \vfill
\end{figure}

%% file: supp/figs/A1.tex
\newlength\gW
\setlength\gW{0.50\linewidth}
\newlength\gH
\setlength\gH{6mm}
\begin{figure}[!h]
     \caption{\textbf{Experiment A-1: Transfer to novel Latin alphabets.} To test generalization to unseen alphabets/novel languages, we test how well our model trained on English transfers to all the other languages in Google1000, namely, French, Spanish and Italian. Below, we show four text-line samples from each language and also similarity map progression as above for one sample.
     These new languages have an expanded alphabet due to accented letters, \eg é, ê, á, û, ù, \etc 
        These are especially challenging as the visual differences between these letters is quite subtle.
     Our model is able to successfully recognize such fine-grained differences and decode these completely unseen languages. For example, in the French similarity map example below, the model successfully distinguishes between `u', `ù' and `û' when their exemplars are provided.}\label{f:a1}
\end{figure}
\vfill
\subsection*{French}
\input{supp/figs/A1_FR}
\simheaderfig{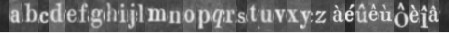}{où plusieurs dentre eux doivent dêjeûner}{où plusieurs dentre eux doivent dêjeûner}
\vfill
\vfill
\vfill
\clearpage

{\setSimWidth{0.3}

\subsection*{Spanish}
\input{supp/figs/A1_ES}

\simheaderfig{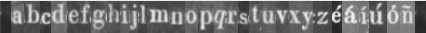}{sabe que la fortuna está subda}{sabe que la fortuna está subda}

\vfill

\subsection*{Italian}
\input{supp/figs/A1_IT}

\simheaderfig{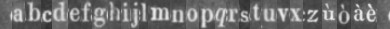}{più facilmente quegli effetti che}{più facilmente quegli effetti che}
}

%% file: supp/figs/A1_FR.tex
\resizebox{\textwidth}{!}{
\begin{tabular}{llll}
  \textbf{\small GT:}   & supportera plus la surtaxe résultant actuellement de la                    & 
  \textbf{\small GT:}   & cane pour léxtradition réciproque des mal                    \\

  \textbf{\small Pred:} & supportera plus la surtaxe résultant actuellement de la                    & 
  \textbf{\small Pred:} & cane pour léxtradition réciproque des mal                    \\

                 & \includegraphics[height=\lheight]{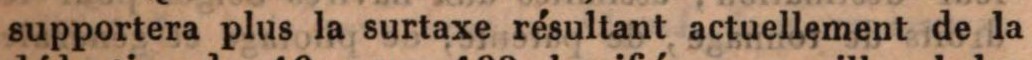}            &
                 & \includegraphics[height=\lheight]{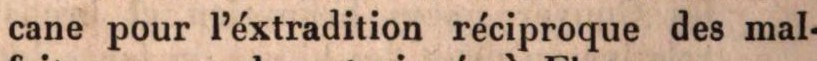}           \\

  \textbf{\small GT:}   & se présente à nos regards de lautre côté des                    & 
  \textbf{\small GT:}   & emploie toutes les forces à approfon                    \\

  \textbf{\small Pred:} & se présente à nos regards de lautre côté des                    & 
  \textbf{\small Pred:} & emploie toutes res forces à approfon                    \\

                  & \includegraphics[height=\lheight]{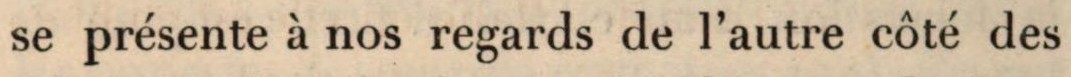}            &
                  & \includegraphics[height=\lheight]{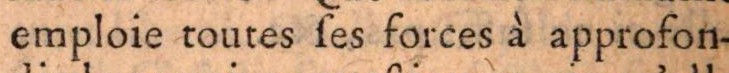}          \\

\end{tabular}

}

%% file: supp/figs/A1_ES.tex
\resizebox{\textwidth}{!}{
\begin{tabular}{llll}
	
  \textbf{\small GT:}   & declaró haber recibido en dote de su segunda mujer                    & 
  \textbf{\small GT:}   & nifestaré con documentos auténticos que tengo en                    \\

  \textbf{\small Pred:} & declaró haber recibido en dote de su segunda mujer                    & 
  \textbf{\small Pred:} & nifestaré con documentos auténticos que tengo en                    \\

                 & \includegraphics[height=\lheight]{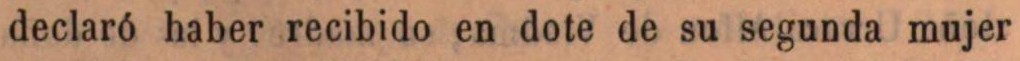}            &
                 & \includegraphics[height=\lheight]{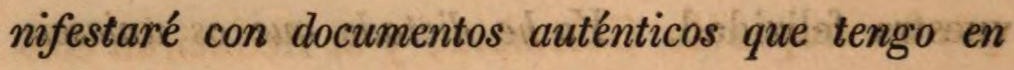}           \\

  \textbf{\small GT:}   & rente á la suerte que le preparaba esa faccion                    & 
  \textbf{\small GT:}   & regresar todos los ausentes y debiendo ser puestos en li                    \\

  \textbf{\small Pred:} & rente á la suerte que le preparaba esa faccion                    & 
  \textbf{\small Pred:} & regresar todos los ausentes y debicndo ber puestos en li                    \\

                  & \includegraphics[height=\lheight]{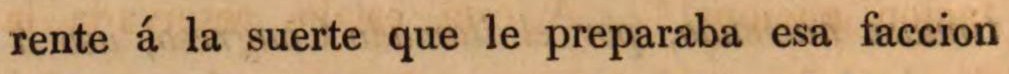}            &
                  & \includegraphics[height=\lheight]{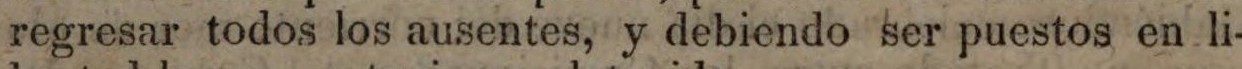}          \\

\end{tabular}

}

%% file: supp/figs/A1_IT.tex
\resizebox{\textwidth}{!}{
\begin{tabular}{llll}

  \textbf{\small GT:}   & \footnotesize{un tal prodigio per indizio di una qualche}                   & 
  \textbf{\small GT:}   & \footnotesize{era già intanto miglior lo stato degli asse}                  \\

  \textbf{\small Pred:} & \footnotesize{un tal prodigio per indizio di una qualche }                    & 
  \textbf{\small Pred:} & \footnotesize{era già intanto miglior lo stato degli asse}                     \\

                 & \includegraphics[height=\lheight]{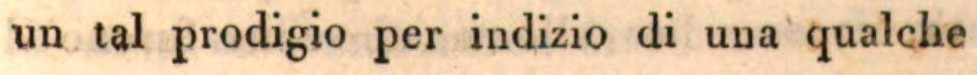}            &
                 & \includegraphics[height=\lheight]{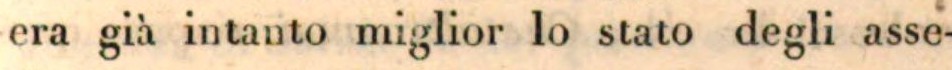}           \\

  \textbf{\small GT:}   & \footnotesize{di cosa che lasciò di essere buona e }                    & 
  \textbf{\small GT:}   & \footnotesize{pio ciò non arrecò meraviglia che a pastori  }                   \\

  \textbf{\small Pred:} & \footnotesize{di cosa che lasciò di essere buona e    }                 & 
  \textbf{\small Pred:} & \footnotesize{piociò nop arrecò meraviglia che a pastori    }                 \\

                  & \includegraphics[height=\lheight]{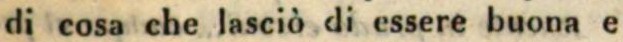}            &
                  & \includegraphics[height=\lheight]{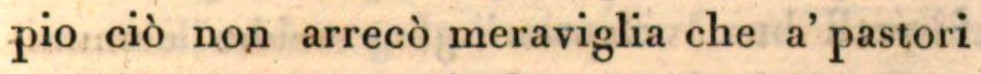}          \\

\end{tabular}

}

%% file: supp/figs/VS4.tex
\begin{figure}[!h]
    \centering
     \caption{\textbf{Experiment A-2: Transfer to non-Latin glyphs.} We evaluate our model on the novel glyph styles in the Omniglot-Sequence test set. The Omniglot-Sequence dataset is constructed by mapping Omniglot glyphs randomly to the English alphabet and then rendering text-line images using those as characters (refer to \Cref{s:exp-data} in the paper). The ground-truth English sentence used for assembling the text-line image is shown underneath the similarity maps. This demonstrates strong generalization to novel visual styles. The position encoder and self-attention module resolve many ambiguities and predict accurate matches.}\label{f:vs4}
     \vspace{6mm}
     \simheaderfig{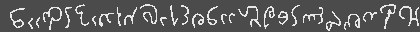}{quite liked the bear}{quite liked the bear}
    \vfill
    
    \simfig{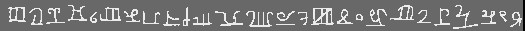}{everybody sideways}{everybody sideways}
    \vfill
    
    \simfig{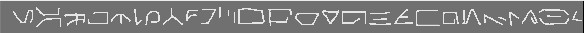}{because its the original}{because its the original}
    \vfill
\end{figure}

%% file: supp/sections/arch.tex
\section{Implementation of SotA models}\label{sec:impl_arch}

In the experiments in the main paper, we compare our model with four state-of-the-art models from three different domains: scene text, handwritten text and document text recognition. In the following subsections, we describe the details of how we implement and use them.

\subsection{Attn. and CTC model -- Baek~\etal~\cite{baek2019wrong}}
We follow the recent work of Baek~\etal~\cite{baek2019wrong} to identify the two of the strongest state-of-the-art text recognition models, named \textit{Attn.} and \textit{CTC} model in the main paper, to benchmark against our method on generalization to novel fonts and alphabets.
They unify modern text recognition methods in a four-stage framework, visualized below -- 
(1)~input transformation, (2)~feature extraction, (3)~sequence modeling and (4)~string prediction.
\input{supp/figs/pipeline.tex}

We do not use the first transformation stage as document line images are mostly rectilinear and are not severely curved/distorted. For the remaining three stages,
we use: (2)~ResNet visual encoder (`Feat.'),
(3)~2-layer bidirectional-LSTM (256 state size) (`Seq.'), and 
for the last `Pred.' stage (4), we consider both: 
(1)~CTC, and (2)~attention based (with a 128-state LSTM) sequence 
prediction. 
The ResNet visual encoder in the baseline models is designed to have
the same (or more) number of parameters as our visual encoder (please refer to \Cref{table:encoder}): baseline ResNet: 6.8M; our encoder: 4.7M parameters. 
Detailed architecture of the visual encoder used in the baseline models is given in the table below.

\vspace{5mm}
\input{supp/tables/base_enc.tex}

\subsection{Industry Standard Model -- Tesseract~\cite{tesseract-ocr}}
Tesseract is a widely-used engine for text recognition in documents like pdf files. They provide free user interface where people can render texts to train the engine on a wide range fonts and  languages. However, 1) the online engine does not support parallel training in gpu, this restricts the batch size to be 1 and training usually runs into divergence , 2) it does not use much data augmentation during training, which leads to serve overfitting problem, 3)  checkpoints are saved by selecting the ones with low training error rates, not from validation results on other fonts. It results in large error rate in generalization to unseen fonts. When the model is trained on our four training attributes in Experiment VS-1, the CER on unseen fonts is 33.5\%. Therefore, we implement Tesseract in PyTorch and train it in the same way as we do for other models.
We follow the standard network architecture they use for English, details are given on Tesseract's github page \footnote{Tesseract's specification of architecture details. \\\url{https://github.com/tensorflow/models/blob/master/research/street/g3doc/vgslspecs.md}}, details are shown in \Cref{table:arch_tess}.
\input{supp/tables/arch_tess.tex}

\subsection{Handwritten Text Model -- Chowdhury~\etal~\cite{chowdhury2018efficient}}

The architecture of the handwritten text recognition model from Chowdhury~\etal~\cite{chowdhury2018efficient} is similar to the \textit{Attn.} model in Baek~\etal~\cite{baek2019wrong}  -- it uses CNNs as the feature extractor, Bi-LSTM as the sequence modelling module and attention-based LSTM as the predictor. However, it adds residual connection~\cite{kim2017residual} and Layer Normalization~\cite{ba2016layer} in their LSTMs.
Codes of this model are not available online. Therefore, we verify our implementation by training and testing on handwriting benchmarks IAM ~\cite{marti2002iam} and RIMES~\cite{augustin2006rimes}. We compare the numbers they report in their paper to ours. Please note that their best performance is achieved with a beam search algorithm in the final LSTM, we do not implement this stage because in our paper we use beam search with an external n-gram language model on every model for a fair comparison.

Our implementation is able to achieve similar or lower CER, but has a consistently higher WER, especially in the IAM  dataset. It might be caused by different methods of measuring WER, \eg the inclusion of punctuations, numbers and abbreviations. Since no details are given in their paper, we treat everything between two white spaces as words when computing the WER, which takes punctuations, numbers as well as words into account.

\input{supp/tables/htr_verify.tex}
\clearpage

%% file: supp/figs/pipeline.tex
 \begin{figure}[!h]
 
  \begin{center}
     \includegraphics[width=\linewidth]{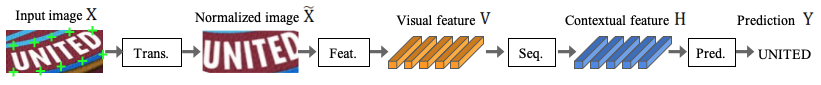}
  \end{center}
     \caption{\textbf{Four stages of modern text-recognition methods.}
     State-of-the-art text recognition models constitute of four stages -- transformation, feature extraction, sequence modelling and prediction. We compare our proposed method against two of the strongest text-recognition models. Figure reproduced from ~\cite{baek2019wrong}}
  \label{f:pipeline}

\end{figure}

%% file: supp/tables/base_enc.tex
\begin{table}[h]
\centering
    \setlength{\tabcolsep}{5pt}
    \renewcommand{\arraystretch}{1.5}
\begin{tabular}{|l|c|c|c|c|c|}
\hline
layer & \textbf{kernel} & \textbf{\begin{tabular}[c]{@{}c@{}}channels\\ in / out\end{tabular}} & \textbf{pool} & \multicolumn{1}{l|}{\textbf{\# conv layers}} & \textbf{\begin{tabular}[c]{@{}c@{}}output size\\ $H{\times}W$\end{tabular}} \\ \hline
conv1 & $3{\times}3$ & 1 / 64 & max = (2, 2) & 4 & $16{\times}W/2$ \\ \hline
resBlock1 & $3{\times}3$ & 64 / 64 & max = (2, 1) x 2 & 6 & $4{\times}W/2$ \\ \hline
resBlock2 & $3{\times}3$ & 64 / 128 & max = (2, 1) x 2 & 6 & $2{\times}W/4$ \\ \hline
resBlock3 & $3{\times}3$ & 128 / 256 & max = (1, 2) & 4 & $2{\times}W/8$ \\ \hline
resBlock4 & $3{\times}3$ & 256 / 512 & max = (1, 1) & 2 & $2{\times}W/8$ \\ \hline
AvgPool & - & - & avg = (2, 1) & - & $1{\times}W/8$ \\ \hline
\end{tabular}

\caption{Architecture details of the visual encoder used for feature extraction.}
\label{tab:base_enc}
\end{table}

%% file: supp/tables/arch_tess.tex
\begin{table}[!htbp]
\centering
	\renewcommand{\arraystretch}{1.2}
	    \setlength{\tabcolsep}{10pt}
	    
		\begin{tabular}{|c|c|c|c|}
			\hline
			layers & kernel & \begin{tabular}[c]{@{}c@{}}channels \\ in / out\end{tabular} & hidden size \\ \hline
			conv & 3x3 & 1 / 16 & -- \\ \hline
			\multicolumn{4}{|c|}{tanh} \\ \hline
			maxpool & 3x3 & 16 / 16 & -- \\ \hline
			LSTM & -- & 16 / 48 & 48 \\ \hline
			Bi-LSTM & -- & 48 / 192 & 96 \\ \hline
			LSTM & -- & 192 / 256 & 256 \\ \hline
			linear & -- & 256 / \#classes & -- \\ \hline
		\end{tabular}%
			\caption{Architecture details of Tesseract. }
		\label{table:arch_tess}

\end{table}

%% file: supp/tables/htr_verify.tex
\setlength{\tabcolsep}{6pt} %
\renewcommand{\arraystretch}{1.2}
\vspace{5mm}

\begin{table}[!htbp]
	\centering
	\resizebox{0.65\textwidth}{!}{
\begin{tabular}{|c|c|cccc|}
	\hline
	\multicolumn{2}{|c|}{} & \multicolumn{2}{c}{IAM} & \multicolumn{2}{c|}{RIMES} \\ \cline{3-6} 
	\multicolumn{2}{|c|}{} & CER & WER & CER & WER \\ \hline
	\multirow{2}{*}{Baseline} & reported & 17.4 & 25.5 & 12.0 & 19.1 \\ \cline{2-6} 
	& implemented &13.5  & 30.9 & 6.2& 16.0\\ \hline
	\multirow{2}{*}{+ LN} & reported & 13.1 & 22.9 & 9.7 & 15.8 \\ \cline{2-6} 
	& implemented & 11.4  & 26.6 & 5.1 & 12.6 \\ \hline
	\multirow{2}{*}{+ LN + Focal Loss} & reported & 11.4 & 21.1 & 7.3 & 13.5 \\ \cline{2-6} 
	& implemented & 10.9  & 25.4 & 4.6 &12.8  \\ \hline
\end{tabular}%
}
	\vspace{3mm}
	\caption{Verification of our implementation of the SotA model~\cite{chowdhury2018efficient} in handwritten text recognition. We compare the error rates they report in their paper with those from our implementation on IAM and RIMES.}
	\label{tab:htr_verify}

\vspace{-6mm}
\end{table}          

%% file: supp/sections/scene.tex
\section{Performance on Scene Text} \label{sec:scene}
Although our model are trained to recognize rectilinear text images, it can be applied to scene text with a transformation module. We use the pre-trained transformation module from Baek~\etal~\cite{baek2019wrong} to transform the images and test our model on commonly used scene text dataset: SVT~\cite{Wang10b}, ICDAR~\cite{Karatzas13,Karatzas15} and IIIT5k~\cite{MishraBMVC12}. Given that the visual style in scene texts images is inconsistent and we don't have access to the fonts, we cross-match the images with exemplars from a randomly selected font. We show the similarity maps and predictions from our model below, with examples of both Arabic numbers and words in various visual styles.

We compare our results with Yao~\etal~\cite{yao2014strokelets} which also learn to recognizes texts by using visual primitives, and show the results in  \Cref{tab:scene-results}.
\input{supp/tables/scene}
\begin{figure}[h]

\caption{\textbf{Visualization of our model's performance on scene text datasets: SVT~\cite{Wang10b}, ICDAR~\cite{Karatzas13,Karatzas15} and IIIT5k~\cite{MishraBMVC12}.} We show the alphabet image, scene text image and the similarity map from the feature encoder for each example. }
\vspace{3mm}

\scenefig{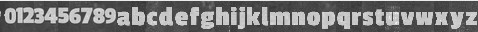}{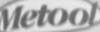}{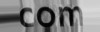}{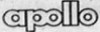}{metool}{com}{apollo}

\scenefig{scene/glyph_line}{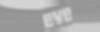}{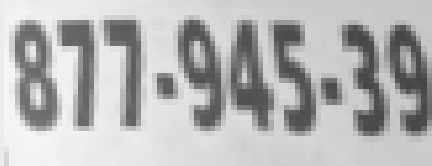}{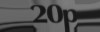}{eve}{87794539}{20p}

\scenefig{scene/glyph_line}{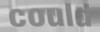}{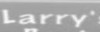}{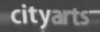}{could}{larry}{cityarts}
\end{figure}

\begin{figure}[h]

\scenefig{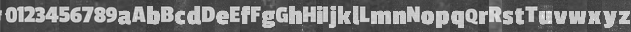}{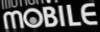}{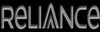}{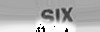}{mobile}{reliance}{six}

\scenefig{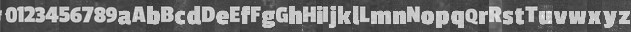}{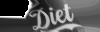}{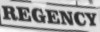}{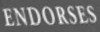}{diet}{regency}{endorses}

\wrongscenefig{scene_upper/upper_glyph_line}{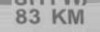}{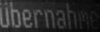}{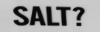}{83km}{ubernahme}{jbernahme}{salt}{salt2}
	
\end{figure}
\clearpage

%% file: supp/tables/scene.tex
\vspace{5mm}
\begin{table}[!htbp]
	\centering
	\renewcommand{\arraystretch}{1.5}
	\setlength{\tabcolsep}{2pt} %
	\resizebox{0.65\textwidth}{!}{%
	\begin{tabular}{cccccc}
		\textbf{Model}                 & \textbf{IIIT5K (small)}           & \textbf{IIIT5K (medium)}          & \textbf{IC03 (full)}              & \textbf{IC03 (50)}                & \textbf{SVT}                      \\ \hline
		{Yao~\etal\cite{yao2014strokelets}} & {80.2}          & {69.3}          & {80.3}          & {88.5}          & {75.9}          \\
		{Ours}       & {\textbf{96.2}} & {\textbf{92.8}} & {\textbf{93.3}} & { \textbf{98.2}} & { \textbf{92.4}} \\ \hline
		\multicolumn{1}{l}{}           & \multicolumn{1}{l}{}              & \multicolumn{1}{l}{}              & \multicolumn{1}{l}{}              & \multicolumn{1}{l}{}              & \multicolumn{1}{l}{}             
	\end{tabular}}
	\caption{Results on scene text dataset IIIT5K, IC03, SVT with lexicon.}
	\label{tab:scene-results}
\end{table}

%% file: supp/sections/dataset.tex
\section{Dataset Details} \label{sec:dataset}

\subsection{Training and Testing Datasets}
We summarize the distribution and size of all our training and testing datasets in \Cref{tab:datasets}.
\input{supp/tables/datasets.tex}

\vspace{5mm}
\subsection{Google1000 Dataset Details}\label{sec:google1k}
We benchmark our method on a large-scale real world dataset of
historical books -- Google1000~\cite{google1000}~(\Cref{s:exp-data}).
Specifically, we employ Google1000 for two experiments:
(1) \textbf{experiment VS-3:} to evaluate our models trained on synthetic data 
for generalization to nuisance factors encountered in the real data like 
degradation, blur, show-through (from behind), inking, fading, oblique text-lines etc
and 
(2) \textbf{experiment A-1:} to evaluate generalization from English (training language)
to novel Latin alphabets/new languages in Google1000, namely French, Spanish and Italian.

For this, we randomly select 40 volumes for each language;
Below, we list the volume-ids for completeness.

\setlength\columnsep{-8pt}
\subsection*{English}
\begin{multicols}{10}
  \begin{enumerate}[label={}]
    \input{supp/scripts/google1k_volumes/en.tex}
  \end{enumerate}
\end{multicols}

\setlength\columnsep{-8pt}
\subsection*{French}
\begin{multicols}{10}
  \begin{enumerate}[label={}]
    \input{supp/scripts/google1k_volumes/fr.tex}
  \end{enumerate}
\end{multicols}

\setlength\columnsep{-8pt}
\subsection*{Spanish}
\begin{multicols}{10}
  \begin{enumerate}[label={}]
    \input{supp/scripts/google1k_volumes/es.tex}
  \end{enumerate}
\end{multicols}

\setlength\columnsep{-8pt}
\subsection*{Italian}
\begin{multicols}{10}
  \begin{enumerate}[label={}]
    \input{supp/scripts/google1k_volumes/it.tex}
  \end{enumerate}
\end{multicols}

\subsection{FontSynth Dataset Details}\label{sec:fontsynth}
\input{supp/figs/font_embed}
The FontSynth dataset (\Cref{s:exp-data})  used for evaluating the effect of increasing
diversity of training fonts on generalization to novel fonts, is created
by taking fonts used in MJSynth~\cite{Jaderberg14c} and splitting them per 
\emph{attributes} as determined from their font names, namely -- 
\emph{regular}, \emph{bold}, \emph{light}, 
\emph{italic}, and \emph{other} (\ie rest of the fonts with none 
of the four attributes in their name).
We select 50 fonts at random from each of the four attributes to create the
training set, while all of the \emph{other} 251 fonts constitute the test set. Data are available to download at:\\
\url{http://www.robots.ox.ac.uk/~vgg/research/FontAdaptor20/}.

\Cref{f:font_embed} visualizes font-embeddings from a font classifier 
trained on the MJSynth fonts. The training font embeddings form 
tight clusters, while the test embeddings are spread out, 
indicating that they are visually distinct from the training set.
Hence, FontSynth forms a good benchmark for evaluating generalization 
to novel fonts.

Please refer to \Cref{f:splits+syn} for a visualization of the fonts in these splits.

%% file: supp/tables/datasets.tex
\setlength{\tabcolsep}{6pt} %
\renewcommand{\arraystretch}{1.2}
\begin{table}[h]
	\centering
	\resizebox{0.6\textwidth}{!}{%
	\begin{tabular}{|c|c|c|c|}
		\hline
		datasets & distribution & number of images & language \\ \hline
		\multicolumn{4}{|c|}{training} \\ \hline
		FontSynth & 200 fonts & 200000 & English \\
		Omniglot-Seq & 30 alphabets $\times$ 20 writers & 60000 & Various \\ \hline
		\multicolumn{4}{|c|}{testing} \\ \hline
		FontSynth & 251 fonts & 12550 & English \\
		Google1000\_EN & 40 books & 2000 & English \\
		Google1000\_FR & 40 books & 2000 & French \\
		Google1000\_ES & 40 books & 2000 & Spanish \\
		Google1000\_IT & 40 books & 2000 & Italian \\
		Omniglot-Seq & 20 alphabets $ \times$ 20 writers & 40000 & Various \\ \hline
\end{tabular}}
\caption{Details of training and testing datasets used in experiments VS1-3 and A1-2.}
\label{tab:datasets}
\end{table}

%% file: supp/scripts/google1k_volumes/en.tex
\item 0194
\item 0612
\item 0050
\item 0287
\item 0701
\item 0315
\item 0047
\item 0131
\item 0343
\item 0034
\item 0093
\item 0174
\item 0025
\item 0084
\item 0676
\item 0380
\item 0006
\item 0128
\item 0448
\item 0059
\item 0100
\item 0513
\item 0347
\item 0151
\item 0663
\item 0441
\item 0511
\item 0340
\item 0255
\item 0709
\item 0335
\item 0068
\item 0147
\item 0224
\item 0218
\item 0176
\item 0054
\item 0587
\item 0438
\item 0569

%% file: supp/scripts/google1k_volumes/fr.tex
\item 0930
\item 0957
\item 0947
\item 0923
\item 0951
\item 0940
\item 0921
\item 0948
\item 0931
\item 0925
\item 0926
\item 0929
\item 0927
\item 0942
\item 0959
\item 0928
\item 0954
\item 0932
\item 0953
\item 0956
\item 0937
\item 0922
\item 0943
\item 0958
\item 0946
\item 0950
\item 0934
\item 0945
\item 0935
\item 0944
\item 0924
\item 0955
\item 0952
\item 0936
\item 0938
\item 0920
\item 0949
\item 0933
\item 0941
\item 0939

%% file: supp/scripts/google1k_volumes/es.tex
\item 0843
\item 0836
\item 0865
\item 0918
\item 0795
\item 0878
\item 0860
\item 0864
\item 0820
\item 0818
\item 0793
\item 0846
\item 0827
\item 0875
\item 0788
\item 0899
\item 0797
\item 0875
\item 0826
\item 0874
\item 0810
\item 0916
\item 0897
\item 0850
\item 0864
\item 0845
\item 0871
\item 0910
\item 0872
\item 0888
\item 0824
\item 0870
\item 0898
\item 0818
\item 0845
\item 0790
\item 0857
\item 0808
\item 0883
\item 0901

%% file: supp/scripts/google1k_volumes/it.tex
\item 0967
\item 0968
\item 0963
\item 0987
\item 0983
\item 0969
\item 0961
\item 0994
\item 0986
\item 0997
\item 0976
\item 0996
\item 0978
\item 0991
\item 0988
\item 0984
\item 0972
\item 0981
\item 0965
\item 0985
\item 0964
\item 0979
\item 0998
\item 0980
\item 0962
\item 0975
\item 0960
\item 0992
\item 0970
\item 0993
\item 0999
\item 0989
\item 0971
\item 0990
\item 0982
\item 0966
\item 0995
\item 0977
\item 0973
\item 0974

%% file: supp/figs/font_embed.tex
\begin{figure}[!h]
  \centering
   \includegraphics[width=0.7\linewidth]{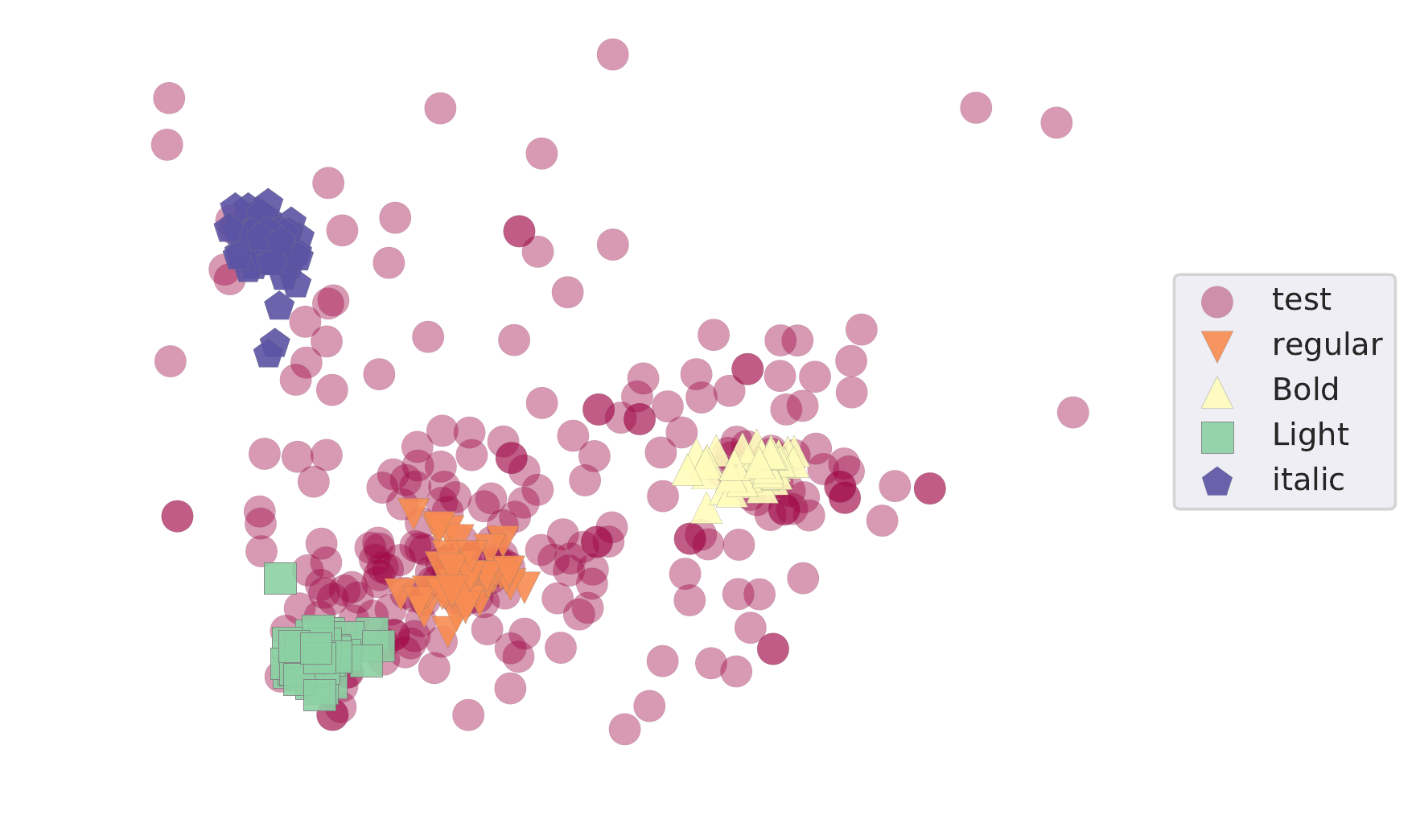}
   \caption{\textbf{FontSynth font embeddings.} The training font splits 
   (\textbf{regular}, \textbf{bold}, \textbf{light}, \textbf{italic}) are distinct from 
   the \textbf{test} fonts.}\label{f:font_embed}
\end{figure}